\documentclass[research]{fcs}

\usepackage{amsmath,amssymb,amsfonts}
\usepackage{algorithmic}
\usepackage{textcomp}
\usepackage{xcolor}

\usepackage{balance}
\usepackage{siunitx}
\usepackage{array}
\usepackage{textcomp}
\usepackage{url}
\usepackage{verbatim}
\usepackage{graphicx}
\usepackage{lipsum}

\usepackage{wasysym}
\usepackage[justification=centering]{caption}
\usepackage[caption=false,font=normalsize,labelfont=sf,textfont=sf]{subfig}
\usepackage{multirow}
\usepackage{diagbox}
\graphicspath{{./figures/}}
\usepackage{array}
\usepackage[ruled,noline,noend]{algorithm2e}

\newtheorem{Definition}{\textbf{Definition}}

\newlength\savewidth
\newcommand\shline{\noalign{\global\savewidth\arrayrulewidth
                            \global\arrayrulewidth 1.0pt}%
                   \hline
                   \noalign{\global\arrayrulewidth\savewidth}
                   }

\title{Gradient Purification: Defense Against Data Poisoning Attack in Decentralized Federated Learning}
\author[1]{Bin Li}
\author[1]{Xiaoye Miao}
\author[1]{Yan Zhang}
\author[2]{Jianwei Yin}
\address[1]{Center for Data Science, Zhejiang University, Hangzhou 310000, China}
\address[2]{College of Computer Science, Zhejiang University, Hangzhou 310000, China}
\fcssetup{
  received       = {month 02, 2025},
  accepted       = {month 07, 2025},
  corr-email     = {miaoxy@zju.edu.cn},
  doi ={10.1007/s11704-025-50240-3}
}

\begin{abstract}
Decentralized federated learning (DFL) is inherently vulnerable to data poisoning attacks, as malicious clients can transmit manipulated gradients to neighboring clients. 
Existing defense methods either reject suspicious gradients per iteration or restart DFL aggregation after excluding all malicious clients.
They all neglect the potential benefits that may exist within contributions from malicious clients.
In this paper, we propose a novel \emph{gradient purification defense}, termed \textsf{GPD}, to defend against data poisoning attacks in DFL.
It aims to separately mitigate the harm in gradients and retain benefits embedded in model weights, thereby enhancing overall model accuracy.
For each benign client in \textsf{GPD}, a recording variable is designed to track historically aggregated gradients from one of its neighbors. 
It allows benign clients to precisely detect malicious neighbors and mitigate all aggregated malicious gradients at once. 
Upon mitigation, benign clients optimize model weights using purified gradients.
This optimization not only retains previously beneficial components from malicious clients but also exploits canonical contributions from benign clients.
We analyze the convergence of \textsf{GPD}, as well as its ability to harvest high accuracy.
Extensive experiments demonstrate that, \textsf{GPD} is capable of mitigating data poisoning attacks under both iid and non-iid data distributions. 
It also significantly outperforms state-of-the-art defense methods in terms of model accuracy.
\end{abstract}

\begin{document}

\section{Introduction}

Federated learning (FL) has emerged as a privacy-preserving computing paradigm, significantly enhancing data utilization across multiple data owners \cite{DBLP:journals/comsur/LimLHJLYNM20, DBLP:journals/kais/LiuHZLJXD22}. 
In FL, multiple clients (i.e., data owners) collaboratively train a global model without transmitting their local training data to others, under the cooperation of the server.
Since the central server is wholesale responsible for distributing and aggregating model updates from clients, any compromise or failure of the server would jeopardize the entire training process \cite{DBLP:journals/pami/KumarMC24}.
As a result, \emph{decentralized} federated learning (DFL) becomes a compelling alternative, wherein each client not only undertakes local training but also functions as an \emph{independent} aggregation server \cite{DBLP:journals/comsur/BeltranPSBBPPC23, DBLP:conf/icml/Shi0WS00T23}. 
DFL does not rely on the central server, thereby circumventing the single-point failures and enhancing robustness \cite{9134967}. 
This paradigm is particularly advantageous for applications in cloud computing environments and financial collaboration \cite{DBLP:journals/dase/RagabSOTPCR24, DBLP:journals/fcsc/SunWTL24, 10.1145/3589334.3645425, DBLP:conf/vldb/000124}.

In the DFL setting, each client transmits its model update (i.e., model weight and gradient) directly to the \emph{neighboring} clients, and \emph{autonomously} aggregates the received updates according to the prescribed \emph{aggregation rule}. 
The aggregated updates are then employed to \emph{refine} each client's local model. Different DFL approaches are characterized by the distinct aggregation rules they adopt. An example is the simple averaging of all received model updates \cite{DBLP:journals/pami/SunLW23}, which enables all participating clients to ultimately converge to a \emph{consistent and optimal} model weight, i.e., transforming from initially disparate local models to a unified global model.

\begin{figure}[t]
    \centering
    \vspace{-10pt}
  \captionsetup[subfloat]{font=footnotesize} 
    \hspace{-20pt}
    \subfloat[\scriptsize The normal DFL]{
        \includegraphics[scale=0.36]{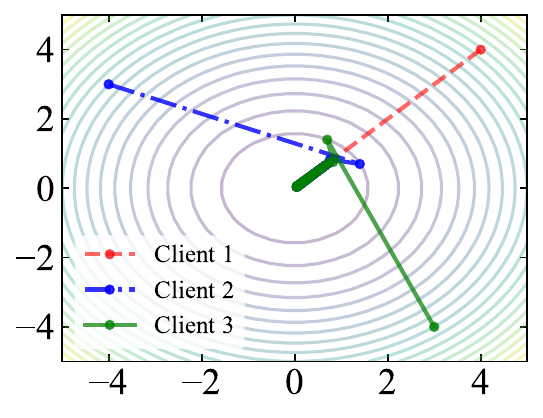}
        \label{Grad_1}
    }
   \hspace{1.00cm} 
    \subfloat[\scriptsize The poisoned DFL]{
        \includegraphics[scale=0.36]{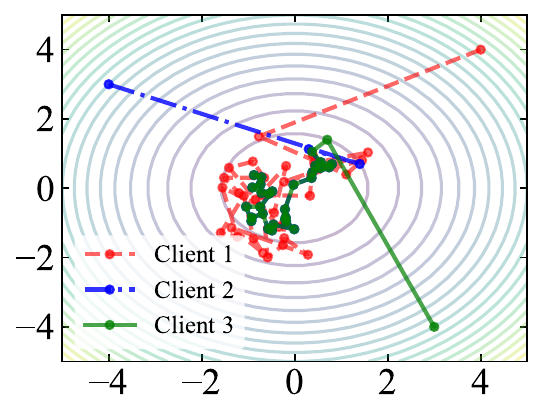}
        \label{Grad_2}
    }
    \caption{Contour of gradient descent in DFL aggregation.}
    \label{Grad}
    \vspace{-10pt}
\end{figure}

However, due to the information exchange among clients, DFL aggregation is particularly susceptible to \emph{data poisoning attacks} \cite{DBLP:journals/fcsc/LuLL24, zheng2024poisoning}. Specifically, the malicious client can manipulate its training data and derive the malicious model gradient. Then this malicious gradient is transmitted to disrupt neighbors' aggregation and ultimately poison the global model.
Fig. \ref{Grad} illustrates the contour of gradient descent across three DFL clients. 
Fig.~\ref{Grad_1} shows that three clients within the normal DFL aggregation ultimately converge to a consistent and optimal state. 
However, when \emph{Client 1} transmits malicious gradient to its neighbors, as shown in Fig. \ref{Grad_2}, the aggregation process between the remaining two clients is significantly disrupted and cannot steadily converge.

Existing defenses against poisoning attacks can be broadly divided into robust methods and detect-restart methods.
\emph{Firstly}, robust methods  \cite{10.1145/3658644.3670307, DBLP:conf/icml/Xie0CL21, DBLP:conf/ndss/RiegerNMS22} aim to \emph{gradually} minimize the impact of potentially malicious clients.
These methods selectively aggregate the median, mean, or a specified range of the received model gradients \emph{per iteration}, to avoid malicious clients as much as possible.
Other emerging methods introduce random noise or regularization per iteration to counteract the malicious impact \cite{DBLP:conf/nips/SunLDHCL21, DBLP:conf/icml/ZhuRC23}. 
However, these methods often accidentally injure benign gradients and ultimately degrade the model accuracy. 
\emph{Secondly}, detect-restart methods \cite{DBLP:conf/ccs/YuanMGYLSWL19, DBLP:conf/esorics/GuptaLND22, DBLP:conf/sp/CaoJZG23} aim to completely mitigate the impact of all malicious clients. 
They primarily detect and exclude malicious clients during a prior aggregation phase, and then restart a \emph{new} DFL aggregation process among remaining benign clients. 
Despite their effectiveness in isolating malicious clients, they rashly discard the potentially beneficial components within the malicious gradients. 
These discarded components could provide critical insights that are absent from all benign clients. 

\begin{figure}
    \centering
    \includegraphics[scale=0.55]{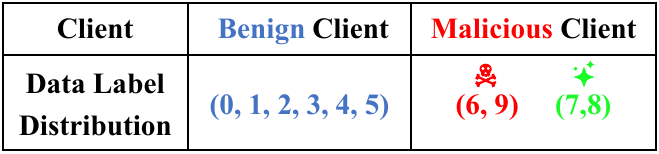}
    \caption{An example of clients' data label distribution.}
    \label{fig:enter-label}
    \vspace{-10pt}
\end{figure}

As illustrated in Fig. \ref{fig:enter-label}, the exclusion of malicious clients effectively mitigates the \emph{misclassified} information pertaining to labels 6 and 9, while completely discarding the \emph{authentic} knowledge associated with true labels 7 and 8. Consequently, if a malicious client possesses the data label that is not present in any benign clients, the restarting aggregation would result in a complete cognitive loss of that data label.

Hence, there are two challenges associated with defending against data poisoning attacks within DFL aggregation.

(i) \emph{Gradual minimization of malicious impact cannot circumvent all malicious clients.}  During the early stage of DFL aggregation, all clients are in disordered states, rendering the distinction between malicious and benign gradients quite subtle. Hence, aggregating the mean or median of received gradients is susceptible to including malicious gradients while inadvertently injuring some benign ones.
Moreover, accurately excluding all malicious gradients per iteration is impractical, malicious gradients are \emph{inevitably} aggregated in certain iterations, ultimately diminishing the model accuracy.

(ii) \emph{Complete mitigation of malicious impact discards the beneficial components from malicious clients.} It is important to note that, there can also be beneficial components within the malicious gradients.
However, existing methods \cite{DBLP:conf/sp/CaoJZG23} utilize the received malicious gradients only to detect and mitigate malicious impact. They neglect the potentially beneficial components within malicious gradients, which are highly likely to enhance model accuracy in heterogeneous data distributions. This is also evidenced by our experimental study.

Therefore, we introduce a novel \emph{gradient purification defense}, termed \textsf{GPD}, in this paper. Inspired by the divide-and-conquer strategy, we employ model gradients and model weights to mitigate the malicious impact and retain beneficial components from malicious clients, respectively.
Each benign client in \textsf{GPD} maintains the \emph{recording variable} that can track historically aggregated gradients from each of its neighbors. 
It enables benign clients to detect the consistency of \emph{historical} gradient records rather than a single gradient submission, thereby precisely identifying malicious neighbors while reserving benign ones. 
It furthermore empowers benign clients to mitigate all aggregated malicious gradients within a single iteration.
Upon mitigation, the previously poisoned model weights are optimized via purified gradients. This optimization leverages the canonical information from benign clients while retaining the beneficial components already contributed by malicious clients, thereby enhancing the model accuracy.
Furthermore, \textsf{GPD} exhibits scalability across a range of detection algorithms and can be effectively integrated with them to identify malicious clients. Our primary contributions can be summarized as follows.

\begin{itemize}
\item We propose the gradient purification defense against data poisoning attacks in DFL, aimed at precisely mitigating malicious impact while retaining benefits to enhance the overall model accuracy.

\item To mitigate the malicious impact, each benign client maintains the recording variable that can track all aggregated gradients from each neighbor. It enables benign clients to detect the historical consistency and mitigate all aggregated malicious gradients at once.

\item To retain beneficial components, the model weights are optimized by purified gradients after mitigation. This optimization skillfully retains the beneficial knowledge previously contributed by malicious clients, and exploits canonical contributions from benign clients. We also analyze the convergence of \textsf{GPD}.

\item  Extensive experiments confirm that, the global model aggregated using \textsf{GPD} harvests the best model accuracy, compared to various state-of-the-art defense methods. It significantly performs better than other baselines in heterogeneous data distributions.
\end{itemize}

The rest of this paper is organized as follows.  Section \ref{section-6} discusses the related work. Section \ref{section-2} provides the background. Section \ref{section-3} introduces the gradient tracking mechanism for the basis of \textsf{GPD}. Section \ref{section-4} elaborates on the \textsf{GPD} solution, including the consistency-based detection algorithm.
Section \ref{section-0} analyzes the performance of \textsf{GPD}. Section \ref{section-5} presents the experimental results and our findings. 
Finally, we conclude this paper in Section \ref{section-7}. 

\section{Related Work}
\label{section-6}
\textbf{Decentralized federated learning (DFL)} aggregation rules can be categorized into two types based on the presence or absence of gradient tracking property, i.e., distributed stochastic gradient descent (DSGD) and distributed stochastic gradient tracking (DSGT). DSGD aggregation rules \cite{10.5555/3295222.3295285,9771388} represent a natural extension of stochastic gradient descent (SGD) in decentralized topology, weighted averaging received model updates and utilizing the double-stochastic matrix to guarantee consistent convergence. However, DSGD aggregation underperforms in heterogeneous data distributions \cite{pmlr-v119-koloskova20a}. DSGT aggregation rules, including \cite{9789732,10.5555/3540261.3541134}, primarily employ an auxiliary variable to track the global gradient (i.e., the average of all local gradients), thus achieving optimal convergence under heterogeneous data distributions \cite{DBLP:conf/nips/AketiH023}.

\textbf{Data poisoning attacks} can be either backdoor attacks or label flipping attacks. 
In backdoor attacks \cite{kumari2023baybfed, DBLP:conf/iclr/XieHCL20}, the malicious client injects a pre-designed trigger into a selected input during training, ensuring that any test input embedded with this trigger will be misclassified. In label flipping attacks \cite{DBLP:conf/icml/BhagojiCMC19}, the malicious client seeks to manipulate the global model such that a test input, specified by the attacker, is misclassified into a target label determined by the attacker.

\textbf{Defenses against poisoning attacks}  can be categorized into robust approaches and detect-restart methods.

The robust aggregation approaches are designed to gradually minimize the potentially malicious impact. The typical methods compute the median of received model updates (i.e., Trimmed Mean \cite{DBLP:conf/icml/YinCRB18}) or filter the outliers in the set of received model updates based on the pairwise distance (i.e., MultiKrum \cite{DBLP:conf/nips/BlanchardMGS17}). Bulyan \cite{DBLP:conf/icml/MhamdiGR18} combines both approaches by first filtering the outliers using MultiKrum and then applying robust aggregation using Trimmed Mean. Byz \cite{10.1145/3658644.3670307} identifies and mitigates malicious gradients via similarity check. However, they accidentally injure benign model updates, thereby eventually declining the overall model accuracy.

The detect-restart methods focus on the advanced detection algorithm to detect malicious clients, and then restart a new DFL aggregation process among benign clients.  For instance, Long-short \cite{DBLP:conf/esorics/GuptaLND22} detects clients via checking clients' model-update consistency.  
FedRecover \cite{DBLP:conf/sp/CaoJZG23} estimates the clients' model updates after excluding all malicious clients. 
However, they miss beneficial contributions from malicious clients and cannot fully activate the data value.

The defense methods highly relevant to this paper about DFL are some \emph{client-self-adaptive} defenses against poisoning attacks, such as FL-WBC \cite{DBLP:conf/nips/SunLDHCL21}, LDP \cite{DBLP:conf/ndss/NaseriHC22}, and LeadFL \cite{DBLP:conf/icml/ZhuRC23}. These methods primarily rely on introducing random noise or employing regularization techniques to implicitly mitigate the harmful impact of malicious clients, while possibly retaining beneficial components from them. 
However, they inevitably impair the benign model updates and finally degrade the model accuracy. Our proposed solution not only precisely mitigates malicious impact but also retains the previously beneficial contributions from malicious clients. It can harvest higher model accuracy than that of these methods.

\section{Preliminaries}
\label{section-2}

\begin{table}[t]
    \centering 
    \caption{\centering The description of the symbols.}
    \vspace{5pt}
    \label{tab:symbol_label}
\resizebox{0.49\textwidth}{!}{%
\Large
\begin{tabular}{cl}
\shline
\textbf{Symbol} & \textbf{Meaning} \\ \hline
$n$ & the number of total clients\\ \hline
$b$ & the number of benign clients\\ \hline
$c_i \in C$ & the set of DFL clients\\ \hline
$\mathcal{D}_i$ & the local private dataset for client $c_i$\\ \hline
$c_j \in \mathcal{N}_i$ &  each of neighbors for client $c_i$ \\ \hline
$|\mathcal{N}_i|$ &  the number of neighboring clients for client $c_i$ \\ \hline
$\mathcal{L}_i(\cdot)$ &the loss function for client $c_i$\\ \hline
$\mathcal{L}_i^M(\cdot)$ &the malicious loss function for malicious client \\ \hline
$T$ &the number of total iterations \\ \hline
$w_{ij}^t$ &the aggregation weight assigned by $c_i$ for  $c_j$ in the $t$-th iteration \\ \hline
$\boldsymbol{\theta}_i^t$ &the local model weight for client $c_i$ in the $t$-th iteration\\ \hline
$\nabla \mathcal{L}_i(\boldsymbol{\theta}_i^t)$ &the local gradient for client $c_i$ in the $t$-th iteration\\ \hline
$\boldsymbol{\gamma}_i^t$ &the local gradient tracking variable for client $c_i$ in the $t$-th iteration \\ \hline
$\boldsymbol{\beta}_{ij}^t$ &the recording variable assigned by client $c_i$ for $c_j$ in the $t$-th iteration \\ \hline
$\lambda$ &the decay step size \\ \shline
\end{tabular}}
\end{table}

In this paper, we consider $n$ clients $C=\left\{c_1, c_2, \cdots, c_n\right\}$ participating in the DFL aggregation, where each client $c_i$ locally holds a private dataset $\mathcal{D}_i$. For illustrative purposes, we do not make specific distributional assumptions about how training data is sampled across clients. For each client $c_i$, let $\mathcal{N}_i$ represent the set of neighboring clients 
(also including $c_i$) that exchange model weights and gradients with client $c_i$. Moreover, client $c_i$ needs to assign an aggregation weight $w_{ij} \in (0,1)$ to each neighbor $c_j \in \mathcal{N}_i$, while this weight is set to 0 for clients with which there is no direct communication.
The symbols defined in this paper are described in Table \ref{tab:symbol_label}. 

\begin{figure}[t]
    \centering
    \vspace{-10pt}
  \captionsetup[subfloat]{font=footnotesize} 
    \subfloat[\scriptsize The normal DFL]{
        \includegraphics[scale = 0.60]{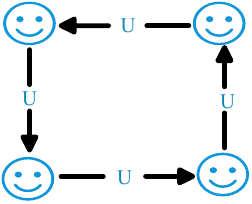}
        \label{DFL_1}
    }
   \hspace{1cm} 
    \subfloat[\scriptsize The poisoned DFL]{
        \includegraphics[scale=0.60]{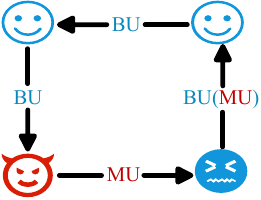}
        \label{DFL_2}
    }
    \caption{The DFL aggregation without/in malicious clients. }
    \label{DFL}
    \vspace{-8pt}
\end{figure}

In general, these $n$ clients collaboratively train the global model to minimize the loss function across all clients. Let $\boldsymbol{\theta}_i$ represent the model weight and $\mathcal{L}_i(\cdot)$ represent the loss function for client $c_i$, the optimization objective function for DFL aggregation can be formulated as follows \cite{nedic2020distributed}.
\begin{align}
\label{obj}
    \begin{aligned}
    \small 
        \mathop{\min} \sum_{i=1}^n \mathcal{L}_i(\boldsymbol{\theta}_i) \quad  s.t.\quad
           \boldsymbol{\theta}_1=\boldsymbol{\theta}_2=\cdots=\boldsymbol{\theta}_n.
    \end{aligned}
\end{align}
In particular, at the $t$-th iteration, as illustrated in Fig. \ref{DFL_1}, where ``U'' represents the model update, each client $c_i$ first sends (resp. receives) model update to (resp. from) each neighbor $c_j \in \mathcal{N}_i$. Subsequently, each client $c_i$ employs the decentralized stochastic gradient descent aggregation (DSGD) \cite{lian2017can} rule to refine its local model. This aggregation form can be expressed as follows.
\begin{equation}
\label{dsgd}
\small 
\boldsymbol{\theta}_i^{t+1} \leftarrow \sum_{j=1}^n w_{ij}^{t}(\boldsymbol{\theta}_j^{t}-\lambda \nabla \mathcal{L}_j(\boldsymbol{\theta}_j^{t})).
\end{equation}
where $w_{ij}^{t}\in(0,1)$ represents the aggregation weight assigned to client $c_j \in \mathcal{N}_i$ by client $c_i$ at the $t$-th iteration, $\lambda$ denotes the decay step size and $\nabla \mathcal{L}_j(\boldsymbol{\theta}_j^{t})$ represents the local model gradient of client $c_j$. Appropriate aggregation weights guarantee that all DFL clients converge to a uniform model weight, i.e., $\boldsymbol{\theta}_1=\boldsymbol{\theta}_2=\dots=\boldsymbol{\theta}_n$, corresponding to the global model.

\textbf{Threat model.}
A malicious client can undermine the DFL aggregation by transmitting the malicious model update to its neighboring clients. 
For instance, as depicted in Fig. \ref{DFL_2}, where ``BU/MU'' denotes the benign/malicious update respectively, a malicious client first transmits the malicious model update to its neighbor. Subsequently, this neighbor, after aggregating the received malicious update, forwards its own \emph{now-poisoned} model update to its other neighbor(s). This cascading propagation of poisons leads to contamination across all clients, ultimately resulting in aggregation failure and reducing the global model accuracy.

Without loss of generality, we assume that malicious clients possess computational and communicational capabilities comparable to those of benign clients. 
They manipulate their training data to transmit malicious gradients that poison the aggregation process. Generally, the malicious client would train its local model to minimize the \emph{malicious loss function} $\mathcal{L}_i^M(\cdot)$ and then inject the malicious gradient into its model update. It can be formulated as follows \cite{DBLP:conf/icml/ZhuRC23}. 
\begin{equation}
\small 
\label{dsgd_m}
\boldsymbol{\theta}_i^{t+1} \leftarrow \sum_{j=1}^n w_{ij}^t\boldsymbol{\theta}_j^{t}-\lambda [(1-\pi) \nabla \mathcal{L}_i(\boldsymbol{\theta}_i^{t}) + \pi \nabla  \mathcal{L}_i^M(\boldsymbol{\theta}_i^{t})].
\end{equation}
The parameter $\pi$ denotes the malicious level.
A higher $\pi$ not only magnifies the malicious impact but also elevates the likelihood of being detected and excluded by benign clients.

\textbf{Existing defenses.} To safeguard against data poisoning attacks within DFL aggregation, existing approaches generally adopt two strategies: either selectively aggregating received model updates, or detecting malicious clients and then restarting the DFL aggregation. In essence, given that there are $b$ benign clients out of $n$ total DFL clients, the primary goal of these defenses is to aggregate a global model equivalent to the global model aggregated exclusively by $b$ benign clients \cite{DBLP:conf/sp/CaoJZG23}, i.e., transform the original objective defined in Eq. \ref{obj} from 
$\mathop{\min} \sum_{i=1}^n \mathcal{L}_i(\boldsymbol{\theta}_i)$ to $\mathop{\min} \sum_{i=1}^b \mathcal{L}_i(\boldsymbol{\theta}_i)$. However, existing methods \emph{neglect} the valuable components within malicious gradients, which minimize the loss more effectively.

\textbf{Our goal.} In this paper, we aim to effectively employ beneficial components from malicious clients while mitigating their harmful impact. Particularly, we observe that model gradients are utilized to optimize model weights, meaning that the useful information within gradients can be captured by the optimized model weights. Therefore, it is feasible to \emph{respectively} mitigate the harmful impact in model gradients and retain benefits embedded in model weights. 
In other words, our goal is to purify gradients and then employ purified gradients to optimize poisoned model weights. 
Notably, this optimization want to retain the malicious client's beneficial components already in poisoned model weights.

\vspace{-1pt}
\begin{Definition} (\emph{Beneficial Component})
The beneficial component is generalizable knowledge relevant to the DFL global learning objective, contributed by the malicious client but absent among all benign clients.
\end{Definition}
\vspace{-3pt}
For instance, if a malicious client is the only one holding a rare data class, the features learned for that class constitute beneficial components. This knowledge is beneficial because it generalizes the global model to a wider data distribution.

When \textsf{GPD} applies the purified gradients to optimize the poisoned model weights, it interacts with the two types of embedded information: i) Beneficial components from malicious client, as generalizable knowledge, align with the main task's optimization landscape. The purified gradients thus tend to preserve or even reinforce these features.
ii) Malicious components from malicious clients, in contrast, are typically brittle, non-generalizable patterns. They are unlikely to align with the purified gradient updates and are therefore progressively ``washed out'' or neutralized by these gradients.

To purify model gradients, we first need to detect malicious clients and then mitigate the malicious impact in model gradients. 
If each benign client records historically aggregated gradients from each of its neighbors, this objective becomes more tractable. Inspired by the gradient tracking mechanism \cite{NEURIPS2022_cd86c6a8} in DFL aggregation, we design the recording variable for benign clients to achieve it, as detailed in Section \ref{section-3}.

\vspace{-1em}
\section{Gradient Tracking Mechanism}
\label{section-3}
Inspired by the gradient tracking mechanism in DFL aggregation, we design a recording variable for benign clients to \emph{track historically aggregated gradients} from each neighbor.
In this section, we first introduce the basic gradient tracking mechanism and then present our adaptation built upon it.

\begin{figure*}[t]
    \centering
  \captionsetup[subfloat]{font=footnotesize} 
    \hspace{-15pt}
    \subfloat[\scriptsize Detect malicious clients and adjust aggregation weight via consistency check]{
        \includegraphics[scale=0.60]{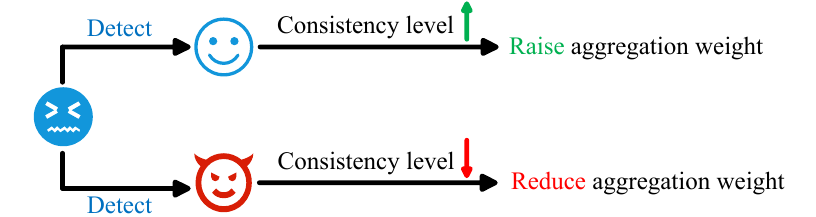}
        \label{over_1}
    }
   \hspace{0.40cm} 
    \subfloat[\scriptsize Mitigate recording variable once aggregation weight is reduced to 0]{
        \includegraphics[scale=0.60]{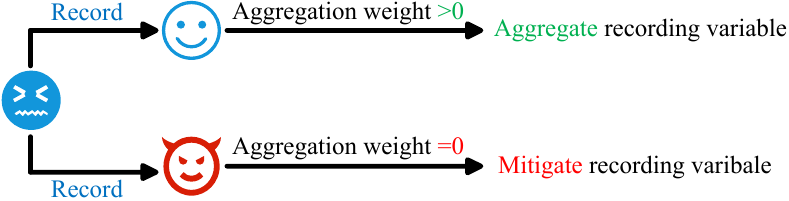}
        \label{over_2}
    }
    \caption{The procedure illustration of gradient purification defense, from the behaviors of one benign client to its neighbors.}
    \label{overview}
    \vspace{-8pt}
\end{figure*}

First, we introduce the gradient tracking mechanism in DFL aggregation, known as the decentralized stochastic gradient tracking (DSGT) aggregation rule. The DSGT aggregation incorporates an auxiliary variable $\boldsymbol{\gamma}$ to track the global gradient (i.e., the average of local model gradients from all clients), thereby accelerating the convergence of DFL aggregation \cite{nedic2020distributed, NEURIPS2022_cd86c6a8}. Specifically, at the $t$-th iteration, after exchanging information with neighbors, each client $c_i$ updates its model weight and auxiliary variable as follows. 
\begin{align}
    \small 
        \boldsymbol{\theta}_i^{t+1} &\leftarrow \sum_{j=1}^n w_{ij}^t (\boldsymbol{\theta}_j^{t}-\lambda \boldsymbol{\gamma}_j^t).  \label{dsgt-w}\\
        \boldsymbol{\gamma}_i^{t+1} &\leftarrow \sum_{j=1}^n w_{ij}^t\boldsymbol{\gamma}_j^{t} + \nabla \mathcal{L}_i(\boldsymbol{\theta}_i^{t+1}) -\nabla \mathcal{L}_i(\boldsymbol{\theta}_i^{t}). \label{dsgt-y}
\end{align}
It can be observed that auxiliary variables are used to optimize the model weights, effectively replacing the original model gradients in DSGD aggregation (i.e., Eq. \ref{dsgd}). Moreover, the aggregation of these auxiliary variables allows them to systematically track the global gradient, i.e., $\sum_{i=1}^n \boldsymbol{\gamma}_i^t = \sum_{i=1}^n \nabla \mathcal{L}_i(\boldsymbol{\theta}_i^{t})$. This implies that auxiliary variables not only tend toward consistency but also constantly track the global gradient, thus guiding the optimization process more effectively. 
For ease of discussion, we will refer to this auxiliary variable $\boldsymbol{\gamma}$ as the \emph{gradient tracking variable} hereinafter.

For instance, Fig. \ref{Grad} illustrates the optimization contour of gradient tracking variables among three clients, where client $c_1$ is alternatively a benign and a malicious client. It is evident that when all three clients are benign, their gradient tracking variables eventually converge to a consistent and optimal state. However, if client $c_1$ transmits its malicious gradient tracking variable, the remaining two benign clients are \emph{unable} to achieve accurate convergence either.
This demonstrates that even a small number of malicious clients can significantly poison the normal DFL aggregation.

In order to defend against data poisoning attacks, we propose an innovative adaptation atop the standard DSGT aggregation. 
Notably, the model weights are mainly optimized using gradient tracking variables, which aggregate both gradient tracking variables from neighbors and their own local model gradients. Therefore, it is sufficient for a benign client to record all historically aggregated gradient tracking variables from each neighbor and subsequently mitigate the malicious portions once detection occurs. We design the recording variable accordingly, which can be defined as follows.

\begin{Definition}(\emph{Recording Variable}) Each benign client $c_i$ in \textsf{GPD} maintains a recording variable $\boldsymbol{\beta}_{ij}^t$ to accumulate historically aggregated gradient tracking variables from its neighbor $c_j$ up to the $t$-th iteration. Let $w_{ij}^t$ denote the aggregation weight assigned to $c_j$ by $c_i$ at the $t$-th iteration, and $\boldsymbol{\gamma}_j^t$ denote gradient tracking variable of $c_j$ at the $t$-th iteration, the recording variable $\boldsymbol{\beta}_{ij}^t$ at the $t$-th iteration can be expressed as:
\begin{align}
\small 
     \boldsymbol{\beta}_{ij}^t \leftarrow \sum_{\tau=0}^t w_{ij}^{\tau} \boldsymbol{\gamma}_j^{\tau}.
\end{align}
\end{Definition}

Leveraging this recording variable, once client $c_j$ is identified as malicious at the $t$-th iteration, client $c_i$ can mitigate all historically aggregated malicious gradients from client $c_j$ at once, by correspondingly setting the $\boldsymbol{\beta}_{ij}^t$ to $\boldsymbol{0}$.

After mitigating the malicious impact in model gradients, the model weights can be optimized by purified gradient aggregation solely among benign clients. This optimization skillfully \emph{retains} the previously aggregated beneficial contributions from malicious clients, and fully employs the canonical contributions from benign clients. Thus, it would significantly enhance the global model accuracy.

It is worthwhile to point out that an important remaining question is how to effectively detect malicious clients. It can be observed that DFL aggregation promotes consistency among clients' gradient tracking variables (as depicted in Fig.~\ref{Grad}).  
Therefore, the benign client can identify a malicious neighbor by examining the discrepancy between its own gradient tracking variable and that of its neighbor.

\newcommand{\sgn}{\mathrm{sgn}}
\begin{algorithm}[t]
\small
\caption{ \centering{The procedure of \textsf{GPD}}}
\LinesNumbered
\SetNlSkip{0.4em}
\label{alg:dsgt} \KwIn{$n$ clients $C=\left\{c_1, c_2, \cdots, c_n\right\}$ including $b$ benign clients, the dataset $\mathcal{D}_i$, the neighbor clients $c_j \in \mathcal{N}_i$ for each client $c_i$, the number of neighboring clients $|\mathcal{N}_i|$, the sign function $\sgn(\cdot)$  outputs 1 when the input is greater than $0.1/|\mathcal{N}_i|$, the decay step size $\lambda$ and the total number of iterations $T$\;}
\KwOut{the consistent and optimal model weight $\boldsymbol{\theta}^{*}$}

\tcc{Initialization}
\For{\emph{each client $c_i\in {C}$ in parallel}}{
set the local model weight $\boldsymbol{\theta}_i^0$ randomly\;
derive the initial gradient $\nabla \mathcal{L}_i(\boldsymbol{\theta}_i^{0})$ based on $\boldsymbol{\theta}_i^0$ and $\mathcal{D}_i$\;
set the initial gradient tracking variable $\boldsymbol{\gamma}_i^0 \leftarrow \nabla \mathcal{L}_i(\boldsymbol{\theta}_i^{0})$ \;
set initial aggregation weight $w_{ij}^0 \leftarrow 1/|\mathcal{N}_i|$ for $c_j$\;
\If{\emph{client $c_i$ is a benign client}}
{
initialize recording variable $\boldsymbol{\beta}_{ij}^{-1} \leftarrow \boldsymbol{0}$ for $c_j$\;
}

}
\tcc{Aggregation}
\For{\emph{iteration} $t=0$ \emph{to} $(T-1)$}{ 
\For{\emph{each client $c_i\in {C}$ in parallel}} {
\If{\emph{client $c_i$ is a benign client}}{
 send $\boldsymbol{\gamma}_i^t$ to neighbor $c_j$, and receive $\boldsymbol{\gamma}_j^t$ from $c_j$ \;
 $\boldsymbol{\beta}_{ij}^t \leftarrow \boldsymbol{\beta}_{ij}^{t-1}+w_{ij}^{t}\boldsymbol{\gamma}_j^{t}$ \;
 \tcp{{\scriptsize \color{gray}update recording variables}}

send $\boldsymbol{\theta}_i^t$ to neighbor $c_j$, and receive $\boldsymbol{\theta}_j^t$ from $c_j$ \;
 $\boldsymbol{\theta}_i^{t+1} \leftarrow \sum_{j=1}^n w_{ij}^t   (\boldsymbol{\theta}_j^t -\lambda   \boldsymbol{\gamma}_j^t)$ \;
  \tcp{{\scriptsize \color{gray}update model weights}}

derive model gradient $\nabla \mathcal{L}_i(\boldsymbol{\theta}_i^{t+1})$  

 $\boldsymbol{\gamma}_i^{t+1} \leftarrow \nabla \mathcal{L}_i(\boldsymbol{\theta}_i^{t+1})  + \sum_{j=1}^n \sgn(w_{ij}^t)\boldsymbol{\beta}_{ij}^t - \sum_{\tau=0}^t \boldsymbol{\gamma}_i^{\tau} $\;
   \tcp{{\scriptsize \color{gray}update gradient tracking variables}}

$w_{ij}^{t+1} \leftarrow \textsf{Detection}(\boldsymbol{\gamma}_j^t, w_{ij}^t)$ for neighbor $c_j$\;
   \tcp{{\scriptsize \color{gray}update aggregation weights}}
}

\If{\emph{client $c_i$ is a malicious client}}{
send $\boldsymbol{\theta}_i^t$, $\boldsymbol{\gamma}_i^t$ and receive $\boldsymbol{\theta}_j^t$, $\boldsymbol{\gamma}_j^t$ with neighbor $c_j$ \;
 $\boldsymbol{\theta}_i^{t+1} \leftarrow \sum_{j=1}^n w_{ij}^0   (\boldsymbol{\theta}_j^t -\lambda   \boldsymbol{\gamma}_j^t)$ \;
derive normal model gradient $\nabla \mathcal{L}_i(\boldsymbol{\theta}_i^{t+1})$ \;
derive malicious model gradient $\nabla \mathcal{L}_i^M(\boldsymbol{\theta}_i^{t+1})$ \;
$\boldsymbol{\gamma}_i^{t+1} \leftarrow (1-\pi) \nabla \mathcal{L}_i(\boldsymbol{\theta}_i^{t+1}) +\pi \nabla  \mathcal{L}_i^M(\boldsymbol{\theta}_i^{t+1})$\;

}}}
\Return $\boldsymbol{\theta}_1^{T} = \cdots = \boldsymbol{\theta}_b^{T} = \boldsymbol{\theta}^{*}$ \;
\end{algorithm}


\section{Solution Overview}
\label{section-4}
In this section, upon the design of recording variables, we first overview our proposed gradient purification defense, i.e., \textsf{GPD} solution. Then, we elaborate on the consistency-based detection algorithm to identify malicious clients.

\subsection{Solution Overview}

As illustrated in Fig. \ref{overview}, \textsf{GPD} primarily involves two steps.
\emph{Firstly}, given that DFL aggregation encourages consistency among clients, we use it to detect malicious clients. 
Specifically, after receiving gradient tracking variables from neighbors, a benign client assesses the consistency level between the received gradient tracking variables and its own ones. 
Based on this consistency assessment, the benign client adjusts the aggregation weight assigned to each of its neighbors. A lower consistency level results in a decreased aggregation weight, effectively reducing the impact of potentially malicious neighbors.
\emph{Secondly}, once a malicious neighbor is identified (i.e., the aggregation weight is reduced to 0), the benign client can swiftly mitigate the recording variable of this malicious neighbor. Since the benign client maintains the recording variable to track historically aggregated gradient tracking variables from its malicious neighbor, it is easy to mitigate all aggregated malicious gradients by setting the corresponding recording variable to zero. Upon mitigation, benign clients would get the purified gradients and subsequently optimize the model weights using these purified gradients.

Algorithm \ref{alg:dsgt} presents the pseudo-code for the \textsf{GPD} solution. The algorithm takes as inputs the set of $n$ clients which includes $b$ benign clients, the local private dataset $\mathcal{D}_i$ for each client $c_i$, the neighboring clients $c_j \in \mathcal{N}_i$ for each client $c_i$, the number of neighboring clients $|\mathcal{N}_i|$, the sign function $\sgn(\cdot)$ which outputs 1 if the input is greater than $0.1/|\mathcal{N}_i|$, the decay step size $\lambda$, and the total number of iterations $T$.
The algorithm outputs the consistent and optimal model weight $\boldsymbol{\theta}^{*}$, corresponding to the global model. 

During initialization, each client $c_i$ begins by randomizing its local model weight $\boldsymbol{\theta}_i^0$ (line 2). 
Following this, client $c_i$ computes the initial model gradient $\nabla\mathcal{L}_i(\boldsymbol{\theta}_i^0)$ based on its local dataset $\mathcal{D}_i$ and the initial model weight $\boldsymbol{\theta}_i^0$ (line 3). 
Subsequently, client $c_i$ sets the initial gradient tracking variable $\boldsymbol{\gamma}_i^0$ equal to its initial gradient $\nabla\mathcal{L}_i(\boldsymbol{\theta}_i^0)$ (line 4). 
Then, for each neighbor $c_j \in \mathcal{N}_i$ and itself, client $c_i$ assigns the initial aggregation weight as $1/|\mathcal{N}_i|$ to equally weigh all neighbors during initialization (line 5). 
Moreover, for benign client $c_i$, the initial recording variable $\boldsymbol{\beta}_{ij}^{-1}$ is set as $\boldsymbol{0}$ (lines 6-7) for each neighbor $c_j$, as the aggregated gradient tracking variable is $\boldsymbol{0}$.

During aggregations \emph{for benign clients}, each client $c_i$ first transmits its gradient tracking variable $\boldsymbol{\gamma}_i^t$ to each neighbor $c_j$, and receives gradient tracking variable $\boldsymbol{\gamma}_j^t$ from each neighbor $c_j$ at the $t$-th iteration (line 11).
Next, benign client $c_i$ updates the recording variable $\boldsymbol{\beta}_{ij}^t$ for each neighbor $c_j$ (line 12). 
Upon exchanging the model weights with neighbors, benign client $c_i$ updates its own model weight, followed by recalculating the corresponding local model gradient based on this newly updated model weight (lines 13–15).
Subsequently, benign client $c_i$ updates its gradient tracking variable, combining the newly calculated model gradient, the recording variables for neighbors, and its own historical gradient tracking variables (line 16). This updated gradient tracking variable will be used to optimize the model weight further.
To detect and exclude malicious neighbors, benign client $c_i$ adjusts the aggregation weight $w_{ij}^{t+1}$ for each neighbor $c_j$ based on the output of the consistency-based detection algorithm (line 17), as described in Algorithm \ref{alg:detect}.

During aggregations \emph{for malicious clients}, these clients exchange model weights and gradient tracking variables with their neighbors following the aggregation rule, just as benign clients do (line 19). They also carry out an honest aggregation of the received model weights and compute the genuine model gradients (lines 20-21). The malicious client does not adjust the aggregation weight, so the aggregation weight is always $w_{ij}^0$. 
However, to poison the global model, malicious clients generate \emph{another malicious gradient} based on specific attack types (line 22).
Finally, malicious clients inject the malicious gradient into the gradient tracking variable, according to their malicious level $\pi$ (line 23).
The key aspect here is that, for malicious clients, doubly manipulating model weight and gradient tracking variable is \emph{avoided}, as these two variables are inherently interdependent. Arbitrary manipulation of \emph{both} could result in inconsistencies, making the malicious behavior easily detectable. 
Instead, malicious clients strategically inject malicious values solely into the gradient tracking variable. This careful manipulation ensures the correspondence between model weight and gradient tracking variable, thereby evading detection by benign clients.

Ideally, benign client $c_i$ continuously reduces the aggregation weight assigned to its malicious neighbors over time. Given that the standard aggregation weight for benign neighbors is established as $1/|\mathcal{N}_i|$ (line 5), we set the detection threshold for malicious neighbors at 1/10 of this normal aggregation weight (i.e., 0.1), to balance sensitivity and robustness.
Once the aggregation weight $w_{ij}$ for malicious neighbor $c_j$ diminishes to $0.1/|\mathcal{N}_i|$ (i.e., a very low aggregation weight), the sign function $\sgn(\cdot)$ outputs 0. At this point, all aggregated malicious gradient tracking variables from neighbor $c_j$ (i.e., the recording variable $\boldsymbol{\beta}_{ij}^t$) will be instantly and completely mitigated by client $c_i$, i.e., $0 \times \boldsymbol{\beta}_{ij}^t$ (line 16).
Upon all mitigation, the subsequent aggregation process will proceed solely among the benign clients, as if malicious clients were never involved, ensuring the aggregation continues smoothly until convergence.
Eventually, \textsf{GPD} returns the consistent and optimal model weight $ \boldsymbol{\theta}_1^{T}= \cdots = \boldsymbol{\theta}_b^{T} = \boldsymbol{\theta}^{*}$, corresponding to the eventual global model for $b$ benign clients.


\begin{algorithm}[t]
\small
\caption{ \centering{The procedure of \textsf{Detection}}}
\LinesNumbered
\label{alg:detect} \KwIn{For client $c_i$, the gradient tracking variable $\boldsymbol{\gamma}_j^t$ received from each neighbor $c_j$, current aggregation weight $w_{ij}^t$ for each neighbor $c_j$\;}
\KwOut{The adjusted aggregation weights $w_{ij}^{t+1}$\;}
\For{\emph{each neighbor $c_j$}}
{
compute the consistency level for $c_j$: \qquad \qquad $s_{ij} \leftarrow e^{-||\boldsymbol{\gamma}_j^t-\boldsymbol{\gamma}_i^t||_2} $ \;
compute the normalized consistency level for $c_j$ $s_{ij}^{\prime} \leftarrow s_{ij}/\sum_{j=1}^n s_{ij}$ \;
adjust the aggregation weight for  $c_j$ $w_{ij}^{t+1} \leftarrow (w_{ij}^t+s_{ij}^{\prime})/\sum_{j=1}^n (w_{ij}^t+s_{ij}^{\prime})$ \;
}
\Return $w_{ij}^{t+1}$ for each neighbor $c_j$\;
\end{algorithm}

\subsection{Consistency-based Detection Algorithm}
In the \textsf{GPD} solution, we primarily leverage the consistency of gradient tracking variable to detect malicious clients, as illustrated in Fig. \ref{over_1}. 
This approach is motivated by the fact that local model gradients from benign clients can differ substantially, especially under heterogeneous data distributions. Therefore, instead of directly comparing these raw, disparate gradients, our consistency-based detection algorithm inspects the gradient tracking variables. As established in prior research \cite{DBLP:conf/nips/AketiH023}, these variables possess a crucial convergence property: irrespective of the data distribution, the variable $\boldsymbol{\gamma}_i^t$ for every benign client $c_i$ eventually converges to a single value—the average gradient across all benign clients.

Algorithm \ref{alg:detect} provides the pseudo-code for the consistency-based detection algorithm.
At the $t$-th iteration, for each benign client $c_i$, it takes as input the gradient tracking variable $\boldsymbol{\gamma}_j^t$ received from each neighbor $c_j$, along with the current aggregation weight $w_{ij}^t$ for $c_j$. 
The algorithm outputs an adjusted aggregation weight $w_{ij}^{t+1}$ for $c_j$.

Firstly, client $c_i$ computes the consistency level $s_{ij}$ between each neighbor $c_j$'s gradient tracking variable and its own one (line 2). 
Specifically, the $s_{ij}$ is calculated using an exponential function of negative Euclidean distance (i.e., L2 norm)  \cite{danielsson1980euclidean} between $\boldsymbol{\gamma_i^t}$ and $\boldsymbol{\gamma_j^t}$.
Following this, client $c_i$ computes the normalized consistency level (line 3).
Subsequently, client $c_i$ adjusts the aggregation weight for each neighbor $c_j$, according to current aggregation weights and normalized consistency levels across all neighbors (line 4).
Finally, the algorithm returns the adjusted aggregation weight $w_{ij}^{t+1}$ (line 5).

Notably, the two normalizations integrate the information from all neighbors and all historical consistency levels (lines 3-4), to adjust the aggregation weight. This provides a robust way for benign clients to manage the impact of each neighbor.
Specifically, it is effective in detecting malicious neighbors that persistently deviate from the prescribed aggregation rule. Moreover, it is tolerant of minor errors, meaning temporarily disrupted benign neighbors are not unjustly excluded, thereby avoiding unnecessary disruption in the DFL aggregation.

It should be emphasized that while we use consistency level to detect malicious clients in this work, the detection algorithm in \textsf{GPD} is highly flexible and can be customized according to specific needs. The primary requirement is to ensure the effectiveness of the detection algorithms and the precision in managing recording variables, thereby ensuring that all aggregated malicious gradients can be accurately identified and mitigated by benign clients.

\section{Performance Analysis}
\label{section-0}
In this section, we analyze the performance of the \textsf{GPD} solution. 
Our primary focus is on the global model accuracy, which is intrinsically \emph{linked to} the convergence of DFL aggregation. 
Hence, the key aspect is proving convergence stability, i.e., ensuring that after mitigating all malicious gradients, the aggregation process within \textsf{GPD} converges in the \emph{same manner} as it would in the absence of malicious clients. This proof is crucial because it states that previously malicious submissions do not disrupt the aggregation convergence, thereby guaranteeing the unaffected model accuracy.

\begin{figure}
    \centering
    \hspace{0pt}
    \includegraphics[scale=0.45]{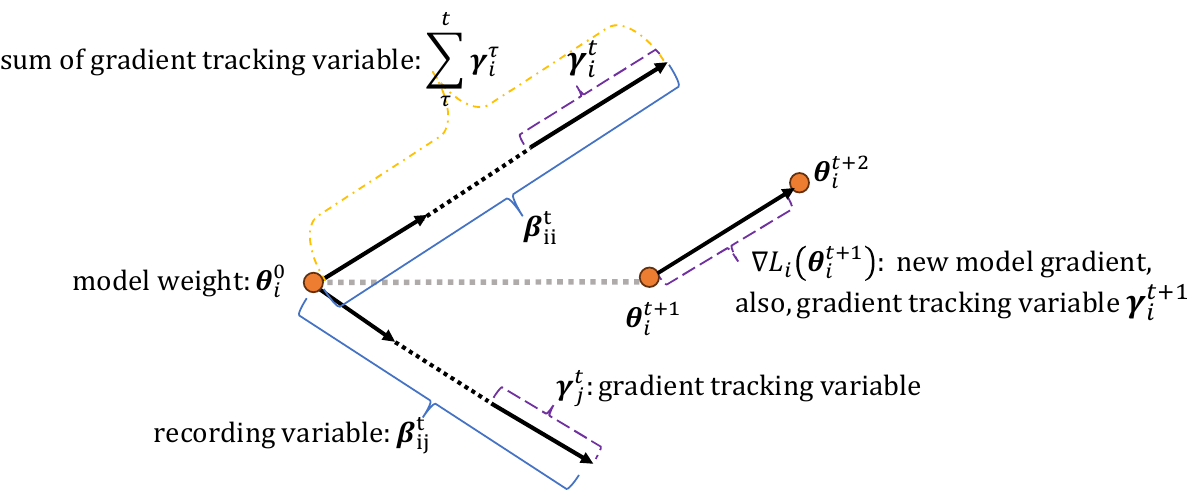}
    \caption{\color{black} The relationships among three variables. If $w_{ij}^t=0$ for identified malicious client $c_j$ by $c_i$, then benign client $c_i$ can do:\\
    (i) $\boldsymbol{\gamma}_i^{t+1}=\nabla \mathcal{L}_i(\boldsymbol{\theta}_i^{t+1})+1\times \boldsymbol{\beta}_{ii}^t+0\times \boldsymbol{\beta}_{ij}^t- \sum_{\tau=0}^t \boldsymbol{\gamma}_i^{\tau} =  \nabla \mathcal{L}_i(\boldsymbol{\theta}_i^{t+1})$\\
    (ii) $\boldsymbol{\theta}_i^{t+2}=1\times(\boldsymbol{\theta}_i^{t+1}-\lambda\boldsymbol{\gamma}_i^{t+1})+0\times(\boldsymbol{\theta}_j^{t+1}-\lambda\boldsymbol{\gamma}_j^{t+1})=\boldsymbol{\theta}_i^{t+1}- \lambda\nabla \mathcal{L}_i(\boldsymbol{\theta}_i^{t+1})$.
}
    \label{Insigh-Update}
    \vspace{-10pt}
\end{figure}

Prior to presenting a rigorous proof, we elucidate the intricate relationships among three essential components: the model weight $\boldsymbol{\theta}$, the gradient tracking variable $\boldsymbol{\gamma}$, and the recording variable $\boldsymbol{\beta}$, to provide an intuitive foundation for understanding \textsf{GPD}'s convergence stability.

As illustrated in Fig. \ref{Insigh-Update}, the recording variable $\boldsymbol{\beta}_{ij}^t$/$\boldsymbol{\beta}_{ii}^t$ systematically documents all aggregated gradient tracking variables from each neighbor $c_j$/$c_i$ \emph{since initialization}. 
When the aggregation weight $w_{ij}^t$ for the malicious client $c_j$ is reduced to 0, the update of $c_i$'s gradient tracking variable $\boldsymbol{\gamma}_i^{t+1}$ effectively mitigates all aggregated malicious gradients from $c_j$, i.e., $0 \times \boldsymbol{\beta}_{ij}^t$.
Furthermore, in the update rule of $\boldsymbol{\gamma}_i^{t+1}$ (line 16 in Algorithm \ref{alg:dsgt}), we subtract the sum of the client's own historical gradient tracking variables. This subtraction is necessary because the recording variables already contain historically accumulated gradients since initialization. To ensure precise global gradient tracking of the \emph{current} gradient, we must subtract the own historical gradient tracking variables as a compensator, represented as $- \sum_{\tau=0}^t \boldsymbol{\gamma}_i^{\tau}$. 
It enables $\boldsymbol{\gamma}_i^{t+1}$ to accurately track the true global gradient, i.e., $\boldsymbol{\gamma}_i^{t+1}=\nabla \mathcal{L}_i(\boldsymbol{\theta_i^{t+1}})$. Subsequently, the model weight $\boldsymbol{\theta}_i^{t+1}$ undergoes optimization using $\boldsymbol{\gamma}_i^{t+1}$ to get $\boldsymbol{\theta}_i^{t+2}$, which effectively reduces to the conventional gradient descent method (considering only one benign client in this simple diagram).

Then, we formulate the convergence theorem and provide its proof about the \textsf{GPD} method. 
In comparison to the standard DSGT aggregation and the aggregation within \textsf{GPD}, the form of model weight aggregation remains identical. Consequently, to establish the convergence stability, it suffices to demonstrate that the aggregation of gradient tracking variable within \textsf{GPD} is the same as standard DSGT aggregation. To demonstrate this, we mainly prove the unaffected global gradient tracking and identical aggregation form of gradient tracking form, as evidenced in Theorem \ref{theo-1}.

\begin{theorem}
\label{theo-1} 
Upon mitigating all recording variables for malicious clients, the subsequent aggregation within \textsf{GPD} ensures that, $\sum_{i=1}^b \boldsymbol{\gamma}_i^t = \sum_{i=1}^b \nabla \mathcal{L}_i(\boldsymbol{\theta}_i^t)$, thereby achieving global gradient tracking. Furthermore, the aggregation form of the gradient tracking variable remains congruent with the standard DSGT aggregation as defined in Eq. \ref{dsgt-y}.
\end{theorem}

\textit{\textbf{Proof: }} We prove Theorem \ref{theo-1} by mainly proving the global gradient tracking property and equivalent aggregation form of gradient tracking variable compared with the standard DSGT.

It is important to preface that all malicious clients can be excluded through a consistency detection \cite{DBLP:conf/esorics/GuptaLND22}.
Once all malicious clients are detected and their corresponding recording variables are completely mitigated (i.e., the aggregation weights in sign function outputs 0), the aggregation of gradient tracking variables changes to the following form.
\begin{align}
\label{out}
\begin{aligned}
\small 
        \boldsymbol{\gamma}_i^{t+1} &=\nabla \mathcal{L}_i(\boldsymbol{\theta}_i^{t+1})  + \sum_{j=1}^n \sgn(w_{ij}^t)\boldsymbol{\beta}_{ij}^t - \sum_{\tau=0}^t \boldsymbol{\gamma}_i^{\tau}  \\
    &=  \nabla \mathcal{L}_i(\boldsymbol{\theta}_i^{t+1}) + \sum_{j=1}^b \boldsymbol{\beta}_{ij}^t - \sum_{\tau=0}^t \boldsymbol{\gamma}_i^{\tau} \\
    &=  \nabla \mathcal{L}_i(\boldsymbol{\theta}_i^{t+1}) + \sum_{\tau=0}^t 
 \sum_{j=1}^b w_{ij}^{\tau} \boldsymbol{\gamma}_j^{\tau}-\sum_{\tau=0}^t 
 \boldsymbol{\gamma}_i^{\tau}.
\end{aligned}
\end{align}
Then, we proceed from Eq. \ref{out} to elucidate the global gradient tracking property of \textsf{GPD} and highlight its equivalent aggregation of gradient tracking variable to the standard DSGT.

\emph{Global Gradient Tracking.} We first prove the global gradient tracking property within the \textsf{GPD}.

Firstly, \textsf{GPD} can guarantee that the aggregation weight matrix composed of elements is a row-stochastic matrix.
Then, We can add up all clients' gradient tracking variables as:
\begin{align}
\begin{aligned}
\small 
        \sum_{i=1}^b \boldsymbol{\gamma}_i^{t+1} &=\sum_{i=1}^b  \nabla \mathcal{L}_i(\boldsymbol{\theta}_i^{t+1}) + \sum_{\tau=0}^t 
\sum_{i=1}^b \sum_{j=1}^b w_{ij}^{\tau} \boldsymbol{\gamma}_j^{\tau}-\sum_{\tau=0}^t \sum_{i=1}^b
 \boldsymbol{\gamma}_i^{\tau} \\
        &=\sum_{i=1}^b  \nabla \mathcal{L}_i(\boldsymbol{\theta}_i^{t+1}) + \sum_{\tau=0}^t  (\sum_{j=1}^b\boldsymbol{\gamma}_j^{\tau}\sum_{i=1}^b w_{ij}^{\tau} - \sum_{i=1}^b  \boldsymbol{\gamma}_i^{\tau}) \\
                &=\sum_{i=1}^b  \nabla \mathcal{L}_i(\boldsymbol{\theta}_i^{t+1}) + \sum_{\tau=0}^t  (\sum_{j=1}^b \boldsymbol{\gamma}_j^{\tau}- \sum_{i=1}^b \boldsymbol{\gamma}_i^{\tau}) \\
                &= \sum_{i=1}^b  \nabla \mathcal{L}_i(\boldsymbol{\theta}_i^{t+1}).
\end{aligned}
\end{align}
Formally, in the aggregation within \textsf{GPD}, the gradient tracking variable $\boldsymbol{\gamma}_i^{0}$ is initialized as $\boldsymbol{\gamma}_i^{0} \leftarrow \nabla \mathcal{L}_i(\boldsymbol{\theta}_i^{0})$. Subsequently, the following identical equation can be obtained.
\begin{align}
\small 
\begin{aligned}
        \sum_{i=1}^b \boldsymbol{\gamma}_i^{t+1} - \sum_{i=1}^b  \nabla \mathcal{L}_i(\boldsymbol{\theta}_i^{t+1}) &= \sum_{i=1}^b \boldsymbol{\gamma}_i^t - \sum_{i=1}^b  \nabla \mathcal{L}_i(\boldsymbol{\theta}_i^t) \\
        &= \cdots \\
        &= \sum_{i=1}^b \boldsymbol{\gamma}_i^0 - \sum_{i=1}^b  \nabla \mathcal{L}_i(\boldsymbol{\theta}_i^0) =\boldsymbol{0}.
\end{aligned}
\end{align}
Therefore, we can affirm that $\sum_{i=1}^b \boldsymbol{\gamma}_i^t = \sum_{i=1}^b \nabla \mathcal{L}_i(\boldsymbol{\theta}_i^t)$, i.e., the \emph{global gradient tracking} within \textsf{GPD} solution.

\emph{Equivalent Gradient Aggregation.}
We aim to show that the aggregation form of gradient tracking variable within \textsf{GPD} is precisely the same as that of standard DSGT aggregation, after mitigating all malicious gradients. In other words, we aim to prove that Eq. \ref{out} equals Eq. \ref{dsgt-y}. Firstly, we can rewrite the Eq. \ref{out} as follows.
\begin{align}
\small 
\begin{aligned}
        \boldsymbol{\gamma}_i^{t+1} =& \nabla \mathcal{L}_i(\boldsymbol{\theta}_i^{t+1}) + \sum_{\tau=0}^t 
 \sum_{j=1}^b w_{ij}^{\tau} \boldsymbol{\gamma}_j^{\tau}-\sum_{\tau=0}^t 
 \boldsymbol{\gamma}_i^{\tau} \\
        =&  \nabla \mathcal{L}_i(\boldsymbol{\theta}_i^{t+1}) + \sum_{\tau=0}^{t-1} \sum_{j=1}^b  w_{ij}^{\tau} \boldsymbol{\gamma}_j^{\tau} +  \sum_{j=1}^b w_{ij}^t \boldsymbol{\gamma}_j^t - \sum_{\tau=0}^{t-1} \boldsymbol{\gamma}_i^{\tau} -\boldsymbol{\gamma}_i^t.\\
\end{aligned}
\end{align}
Because at the $t$-th iteration, the aggregation form of gradient tracking variable is as follows.
\begin{align}
\small 
\begin{aligned}
        \boldsymbol{\gamma}_i^{t} =& \nabla \mathcal{L}_i(\boldsymbol{\theta}_i^t) + \sum_{\tau=0}^{t-1} 
 \sum_{j=1}^b w_{ij}^{\tau} \boldsymbol{\gamma}_j^{\tau}-\sum_{\tau=0}^{t-1} 
 \boldsymbol{\gamma}_i^{\tau}. 
\end{aligned}
\end{align}
Then we can prove the \emph{Eq. \ref{out} = Eq. \ref{dsgt-y}} as follows.
\begin{align}
\small 
\begin{aligned}
        \boldsymbol{\gamma}_i^{t+1} =&  \nabla \mathcal{L}_i(\boldsymbol{\theta}_i^{t+1}) + \sum_{\tau=0}^{t-1} \sum_{j=1}^b  w_{ij}^{\tau} \boldsymbol{\gamma}_j^{\tau} +  \sum_{j=1}^b w_{ij}^t  \boldsymbol{\gamma}_j^t -\sum_{\tau=0}^{t-1} \boldsymbol{\gamma}_i^{\tau} -\boldsymbol{\gamma}_i^t\\
         =& \nabla \mathcal{L}_i(\boldsymbol{\theta}_i^{t+1}) + \boldsymbol{\gamma}_i^t - \nabla \mathcal{L}_i(\boldsymbol{\theta}_i^t) +  \sum_{j=1}^b w_{ij}^t \boldsymbol{\gamma}_j^t - \boldsymbol{\gamma}_i^t \\
         =& \nabla \mathcal{L}_i(\boldsymbol{\theta}_i^{t+1}) - \nabla \mathcal{L}_i(\boldsymbol{\theta}_i^t) +  \sum_{j=1}^b w_{ij}^t \boldsymbol{\gamma}_j^t. 
\end{aligned}
\end{align}

Based on the above analysis, we conclude that the aggregation form of gradient tracking variable within \textsf{GPD} and that of standard DSGT aggregation are fundamentally equivalent. Moreover, the global gradient tracking among benign clients is not impacted by previously malicious gradients. 
Then we can guarantee the purely unaffected gradient aggregation.

Next, we extend our analysis to the aggregation of model weights. 
Although model weights are optimized by previously malicious gradient tracking variables and thus are poisoned. After detecting and mitigating all malicious gradients, the model weights can be effectively corrected through subsequently purified gradients from benign clients. 
This adjustment is analogous to a shift in the initial random point of aggregation process; provided that subsequent gradients are accurate, it can still be guaranteed to converge to optima.
$\hfill\square$ 

\begin{figure}[t]
    \centering
    \vspace{-10pt}
  \captionsetup[subfloat]{font=footnotesize} 
    \hspace{-2.0em} 
    \subfloat[\scriptsize Optimization from original weights]{
        \includegraphics[scale=0.50]{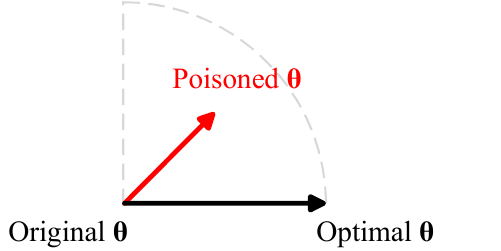}
        \label{Optim_1}
    }
   \hspace{-0.8em} 
    \subfloat[\scriptsize Optimization from poisoned weights]{
        \includegraphics[scale=0.50]{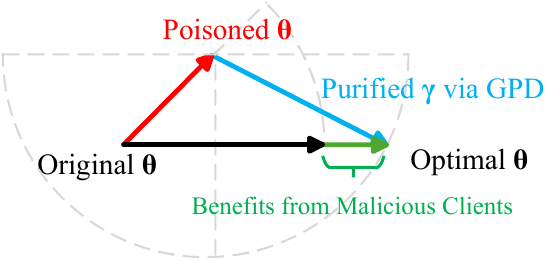}
        \label{Optim_2}
    }
    
    \caption{The aggregation convergence from different starting points.}
    \label{Optim}
    \vspace{-8pt}
\end{figure}

\emph{Notably}, as illustrated in Fig. \ref{Optim_1}, malicious gradients can initially poison model weights and divert them from their intended convergence path. In an ideal scenario, optimization could proceed from original model weights without interference from poisoning attacks.
\emph{Intriguingly}, malicious clients' gradients often contain valuable components that may be unknown to benign clients. As demonstrated in Fig. \ref{Optim_2}, \textsf{GPD} can achieve a \emph{superior} optimal solution compared to the original one. This enhancement occurs because beneficial components previously contributed by malicious clients remain embedded in the model weights, effectively expanding the optimization scope (represented by the rightward shift in the diagram) and consequently improving global model accuracy.

\section{Experimental Evaluation}
\label{section-5}

In this section, we evaluate the performance of our proposed solution and other baselines.
All algorithms were implemented in Python. The experiments were conducted on an Intel Core 2.80GHz server with TITAN Xp 12GiB (GPU) and 192GB RAM, running the Ubuntu 18.04 system.

\textbf{Datasets.} In the experiments, we utilize four public real-world datasets. 
(i) The \textit{MNIST} dataset is a widely known benchmark hand-written digit dataset that includes 70,000 grayscale images of handwritten digits, each 28x28 pixels.
(ii) The \textit{FashionMNIST} dataset is another benchmark dataset in the field of computer vision. 
It includes 70,000 grayscale images of fashion items (where each item is represented by numbers from 0 to 9), each 28x28 pixels. 
(iii) The \textit{CIFAR-10} dataset is a respected benchmark dataset in the realm of object detection and recognition. It includes 60,000 color images categorized into 10 different classes that represent real-world objects (like horses, cats, and airplanes). Each image measures 32x32 pixels within 3 channels. 
(iv)  The \textit{20Newsgroups} dataset is a classic benchmark in text classification. It consists of approximately 20,000 newsgroup documents, partitioned nearly evenly across 20 distinct newsgroups, each corresponding to a different topic or subject area (such as sports, politics, science, and technology).

\textbf{Baselines.} We primarily compare seven defense methods against data poisoning attacks in DFL aggregation. All methods uniformly employ the consistency-based detection algorithm presented in this paper.
\begin{itemize}
\item \emph{Upper} represents the upper bound for detect-restart methods, i.e., the standard DSGT aggregation performed solely by $b$ benign clients. \emph{Only} when all malicious clients are accurately detected and excluded can the global model of restarted aggregation achieve the same accuracy as Upper method.
\item \emph{LDP} \cite{DBLP:conf/ndss/NaseriHC22} counteracts the impact of malicious gradients by incorporating random noise into the aggregated model gradients. However, it inevitably injures benign gradients, thereby degrading the model accuracy.
\item \emph{WBC} \cite{DBLP:conf/nips/SunLDHCL21} selectively adds noise to the parameter dimensions most affected by malicious gradients, thereby minimizing the negative impact on the model accuracy.
\item \emph{Lead} \cite{DBLP:conf/icml/ZhuRC23} introduces regularization to mitigate the impact of malicious gradients while maintaining the model accuracy as much as possible.
\item \emph{Byz} \cite{10.1145/3658644.3670307} identifies and mitigates malicious gradients via similarity check per iteration.
\item \emph{Lower} aggregates all received gradients without any defenses, denoting a reference to evaluate various data poisoning attacks.
\item \textsf{GPD} denotes our gradient purification defense.
\end{itemize}

We do not include the server-side defense such as FedRecover \cite{DBLP:conf/sp/CaoJZG23} in our comparisons, as they necessitate the \emph{central server} to estimate the model updates from all clients. It is unrealistic for each DFL client to estimate all neighbors' model updates per iteration due to the resource limitations and topology invisibility in DFL. What's more, \emph{even} with perfect estimation capabilities and unlimited resources, FedRecover's performance would still be bounded by the Upper method (aggregation of only benign clients). As such, we have already included experimental results for the Upper method.
\begin{figure*}[t]
\centering
\vspace{-23pt}
    \begin{minipage}{0.98\textwidth} 
    \captionsetup[subfloat]{font=footnotesize} 
    \subfloat[\scriptsize iid]{
            \includegraphics[scale=0.40]{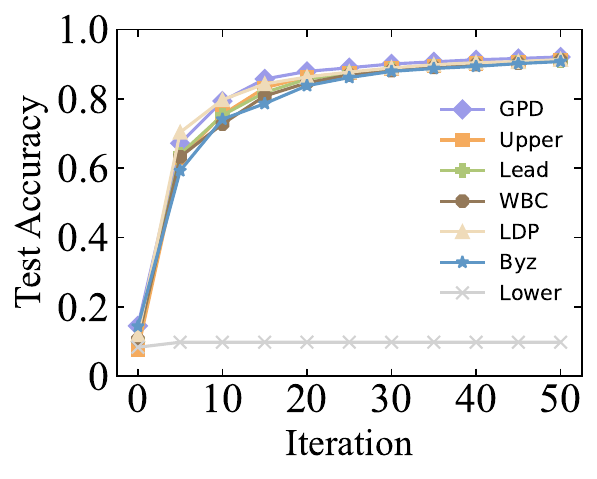}
        }
        \subfloat[\scriptsize non-overlap]{
            \includegraphics[scale=0.40]{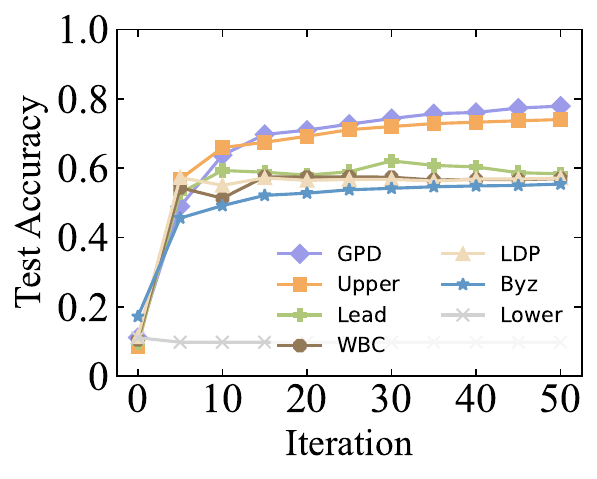}
        }
        \subfloat[\scriptsize label-dir]{
            \includegraphics[scale=0.40]{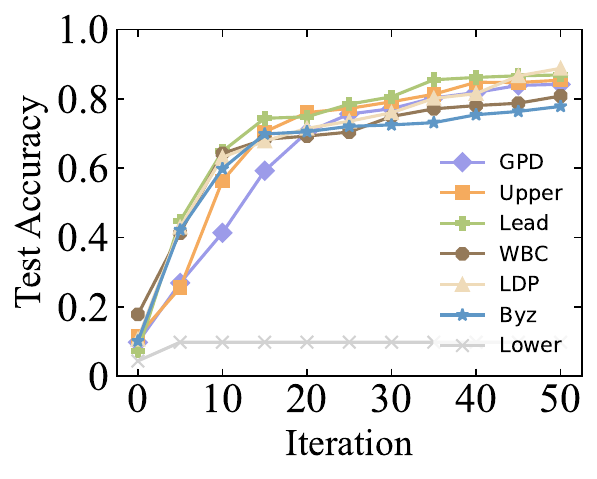}
        }
        \subfloat[\scriptsize quantity-dir]{
            \includegraphics[scale=0.40]{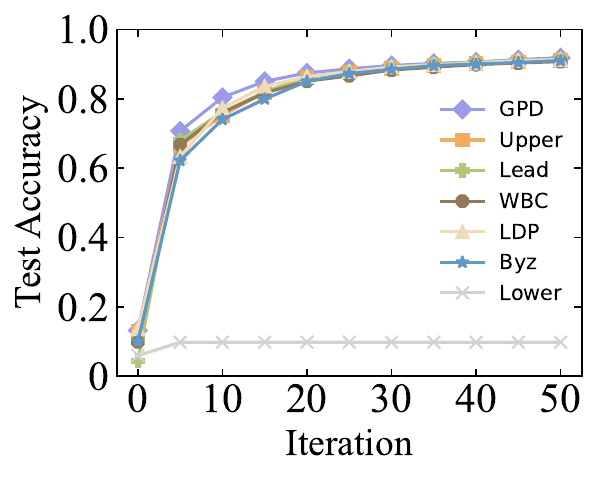}
        }
        \vspace{-3pt}
    \caption{The test accuracy among seven defenses and four data distributions under \emph{Backdoor-9-pixel} attack.}
                    \vspace{-5pt}
    \label{back-9-acc}
    \end{minipage}

    \begin{minipage}{0.98\textwidth} 
    \captionsetup[subfloat]{font=footnotesize} 
    \subfloat[\scriptsize iid]{
            \includegraphics[scale=0.40]{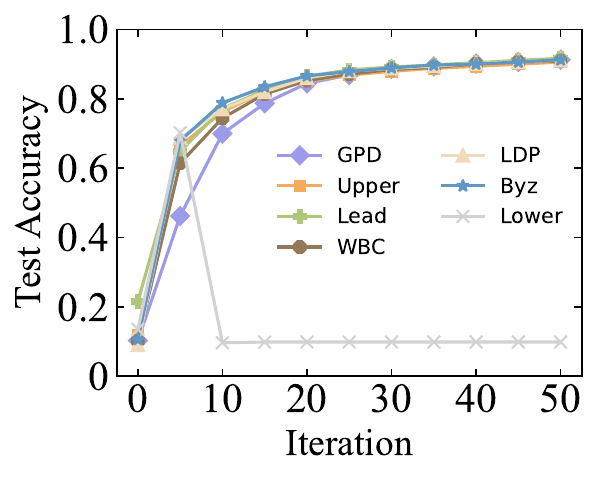}
        }
        \subfloat[\scriptsize non-overlap]{
            \includegraphics[scale=0.40]{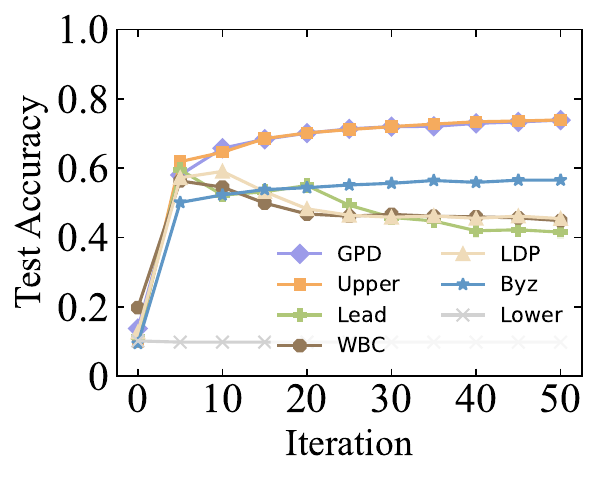}
        }
        \subfloat[\scriptsize label-dir]{
            \includegraphics[scale=0.40]{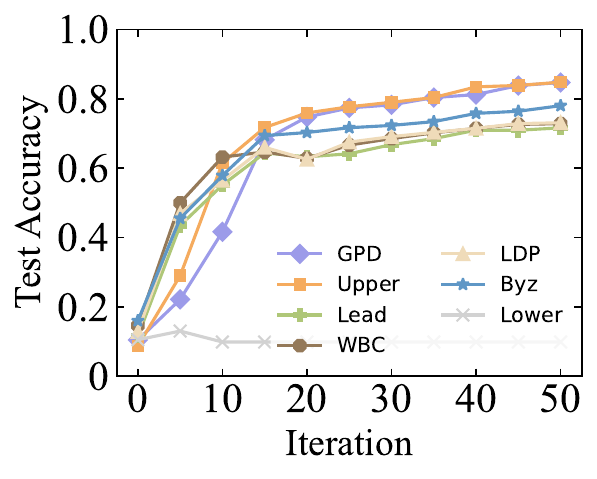}
        }
        \subfloat[\scriptsize quantity-dir]{
            \includegraphics[scale=0.40]{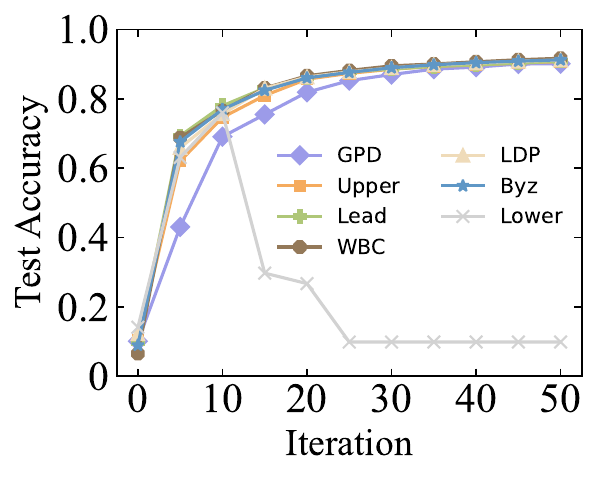}
        }
        \vspace{-3pt}
    \caption{The test accuracy among seven defenses and four data distributions under \emph{Lie} attack.}
        \vspace{-5pt}
    \label{lie-acc}
    \end{minipage}

        \begin{minipage}{0.98\textwidth} 
        \hspace{5pt}
        \captionsetup[subfloat]{font=footnotesize} 
        \subfloat[\scriptsize iid]{
            \includegraphics[scale=0.41]{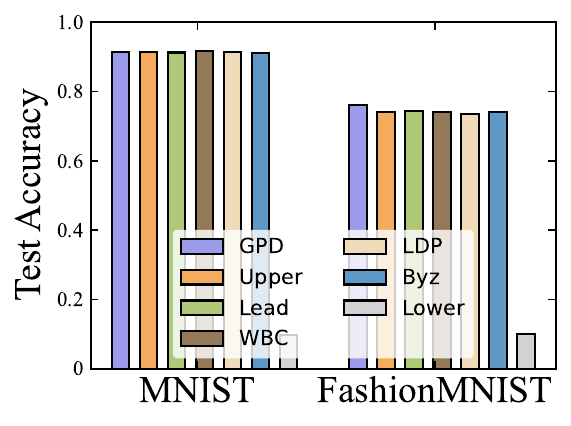}
        }
        \subfloat[\scriptsize non-overlap]{
            \includegraphics[scale=0.41]{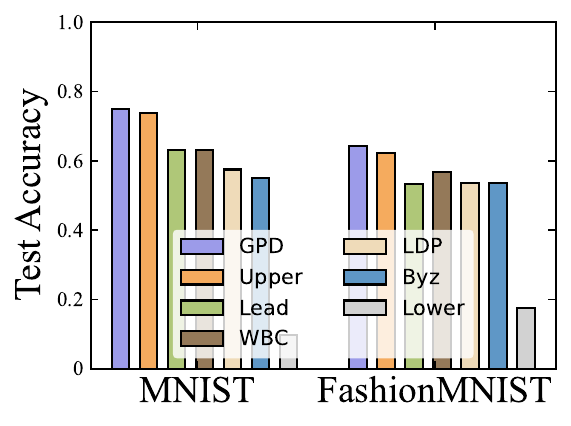}
        }
        \subfloat[\scriptsize label-dir]{
            \includegraphics[scale=0.41]{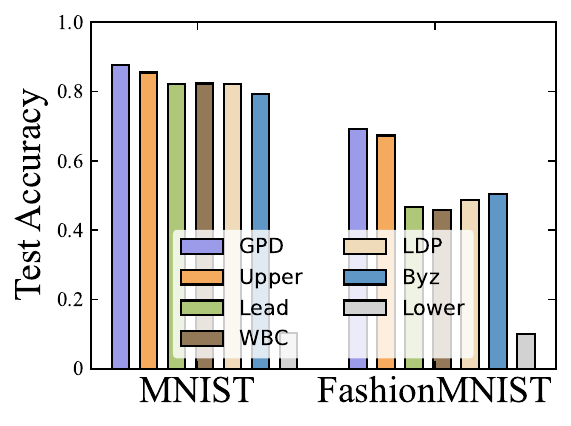}
        }
        \subfloat[\scriptsize quantity-dir]{
            \includegraphics[scale=0.41]{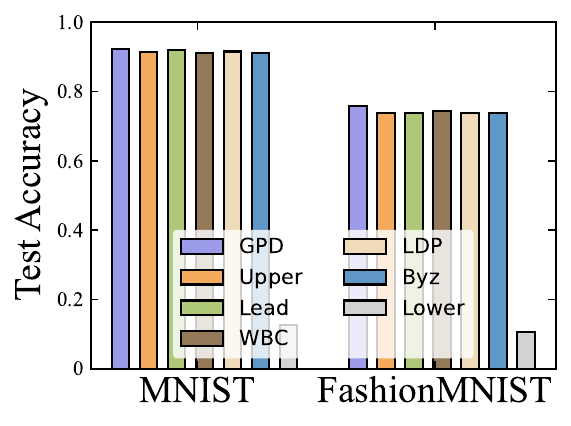}
        }
                \vspace{-3pt}
    \caption{\color{black} The average accuracy among six poisoning attacks under the \textit{MNIST} and \textit{FashionMNIST} dataset.}
    \label{data-1}
    \end{minipage}

            \begin{minipage}{0.98\textwidth} 
        \hspace{5pt}
        \captionsetup[subfloat]{font=footnotesize} 
        \subfloat[\scriptsize iid]{
            \includegraphics[scale=0.41]{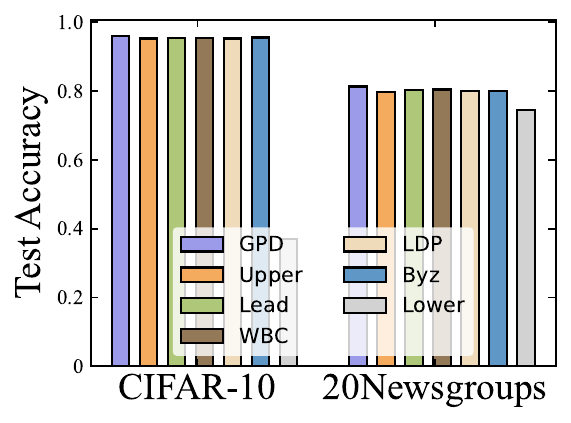}
        }
        \subfloat[\scriptsize non-overlap]{
            \includegraphics[scale=0.41]{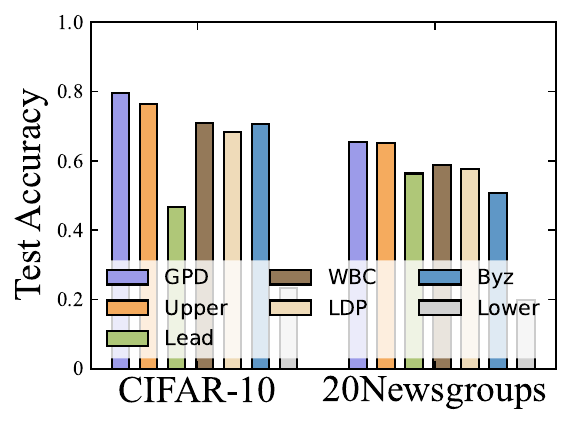}
        }
        \subfloat[\scriptsize label-dir]{
            \includegraphics[scale=0.41]{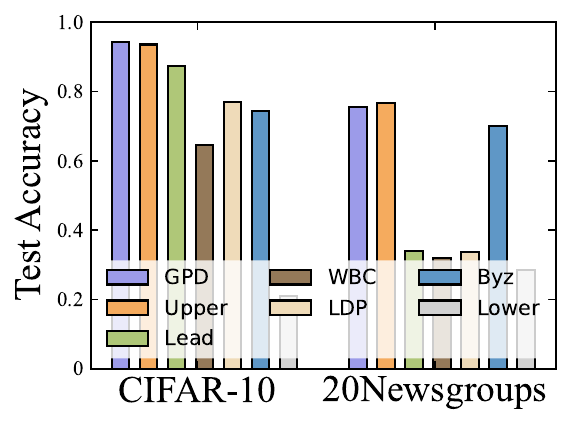}
        }
        \subfloat[\scriptsize quantity-dir]{
            \includegraphics[scale=0.41]{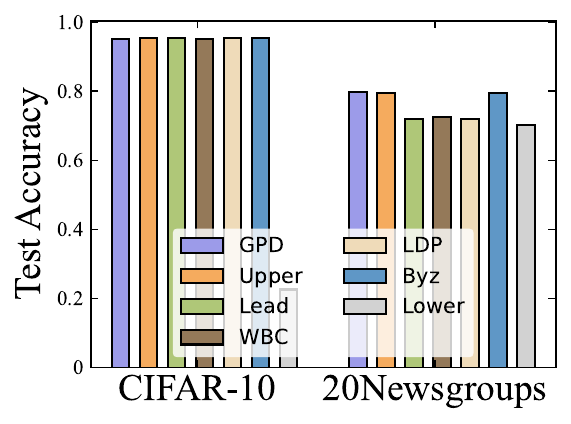}
        }
                \vspace{-3pt}
    \caption{The average accuracy among six poisoning attacks under the \textit{CIFAR-10} and \textit{20Newsgroups} dataset.}
    \label{data-2}
    \end{minipage}
    \vspace{-8pt}
\end{figure*}

\textbf{Data poisoning attacks setting.} We assess both backdoor attacks and label flipping attacks for evaluating various data poisoning attacks. 
\begin{itemize}
    \item \emph{Backdoor-9-pixel} attack \cite{bagdasaryan2021blind} manipulates global model by embedding 9-pixel pattern into training images.
    \item \emph{Backdoor-1-pixel} attack \cite{bagdasaryan2021blind} manipulates global model by embedding the 1-pixel trigger into training images.
    \item \emph{Single-image} attack \cite{DBLP:conf/icml/BhagojiCMC19}  incorporates an identical misclassified image into the training dataset.
\end{itemize}

We also evaluate more powerful attacks including \emph{Lie} attack \cite{DBLP:conf/nips/BaruchBG19}, \emph{Fang} attack \cite{DBLP:conf/uss/FangCJG20}, and \emph{Ndss} attack \cite{DBLP:conf/ndss/ShejwalkarH21}, where the attackers can access all model parameters of benign clients to craft sophisticated poisoning attacks.

\textbf{Data distributions setting.}  We use both independent and identical data distribution (i.e., iid), and non-iid data distribution in our experiments.  In the iid setting, data samples are uniformly distributed to each client.  In the non-iid setting, we deploy the label distribution skew and quantity distribution skew \cite{li2022federated}. We do not use the feature distribution skew where diverse noise is added to the raw data, which would interfere with \emph{LDP} and \emph{WBC} methods. For label distribution skew, we consider non-overlap distributions (i.e., non-overlap) and Dirichlet distribution (i.e., label-dir).
Non-overlap implies that the data label for each client does not intersect, i.e., \emph{a single data label is assigned to only one client}. Label-dir indicates that the data label for each client is distributed following the Dirichlet distribution. In the quantity distribution skew, the data quantity per client is also distributed according to the Dirichlet distribution (i.e., quantity-dir).

\textbf{Implementation details.}
In the default experimental setup, ten clients within the fully-connected communication topology were configured. The decay step size is by default set to 0.01, and the batch size is configured at 256. The malicious label is by default to 1.0 and the ratio of malicious clients to all clients is set to 0.2 by default. 
The scale parameter of the Dirichlet distribution is set as 0.1 by default (i.e., extremely skewed distribution).
A total of 50 iterations is set to enable each method to achieve a stable state, with the local epoch defaulting to 1. Laplace noise with a scale parameter of 0.0001 is employed for \emph{LDP} and \emph{WBC} methods.
Moreover, we utilized a convolutional neural network (CNN) for the \textit{MNIST} and \textit{FashionMNIST} datasets, a Vision Transformer (ViT) for the \textit{CIFAR-10} dataset, and a multilayer perceptron (MLP) for the \textit{20Newsgroups} dataset. While ResNet-18 and Swin Transformer models were also evaluated on \textit{CIFAR-10}, ViT demonstrated superior performance.

\textbf{Metrics.} Our objective is to maximize the model accuracy on the test dataset while simultaneously mitigating the malicious impact of data poisoning attacks. Thus, we primarily consider two metrics in this paper:
\begin{itemize}
    \item \emph{Test Accuracy} represents the proportion of images correctly classified by the global model on the test dataset that does not contain triggers or alterations. Higher test accuracy indicates stronger defense performance.
    \item \emph{Attack Accuracy} represents the proportion of images wrongly classified by the global model (i.e., attack success rate, ASR), which quantifies how successful the attacker is in embedding a backdoor or inverse label into the global model under data poisoning attacks. Lower attack accuracy indicates stronger defense performance.
\end{itemize}

\begin{figure*}[t]
\centering
\vspace{-10pt}
\hspace{-5pt}
    \begin{minipage}{0.98\textwidth} 
    \captionsetup[subfloat]{font=footnotesize} 
    \subfloat[\scriptsize iid]{
            \includegraphics[scale=0.40]{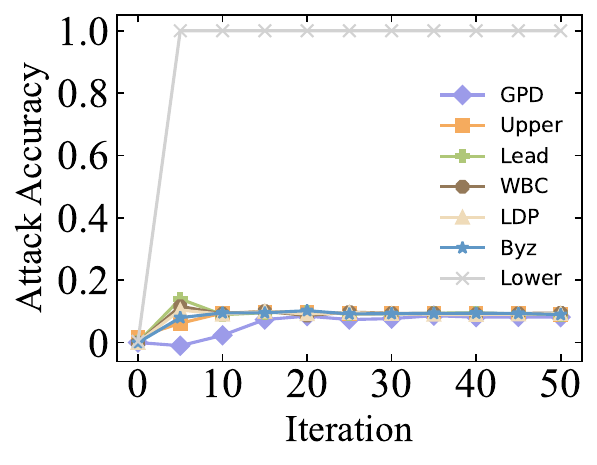}
        }
        \subfloat[\scriptsize non-overlap]{
            \includegraphics[scale=0.40]{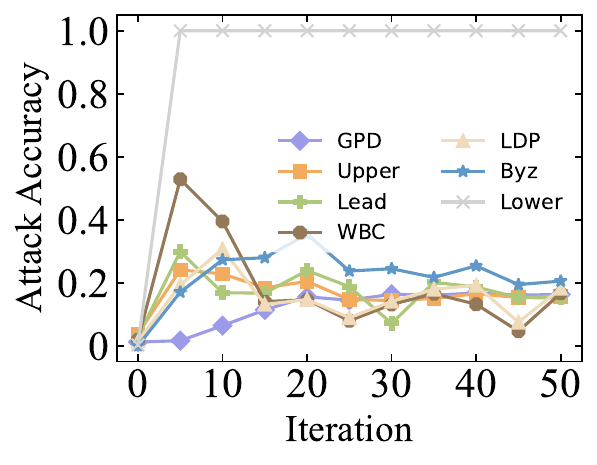}
        }
        \subfloat[\scriptsize label-dir]{
            \includegraphics[scale=0.40]{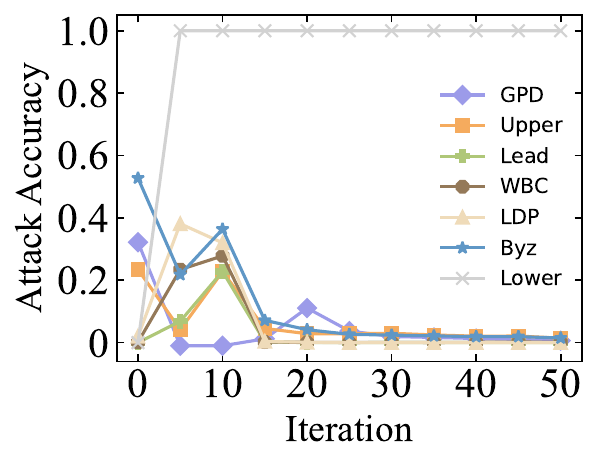}
        }
        \subfloat[\scriptsize quantity-dir]{
            \includegraphics[scale=0.40]{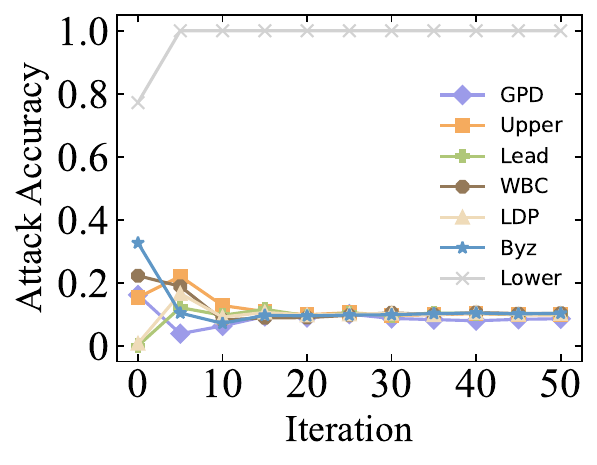}
        }
                        \vspace{-3pt}
    \caption{The attack accuracy among seven defenses and four data distributions under \emph{Backdoor-9-pixel} attack.}
                    \vspace{-5pt}
    \label{back-9}
    \end{minipage}

    \begin{minipage}{0.98\textwidth} 
    \captionsetup[subfloat]{font=footnotesize} 
    \subfloat[\scriptsize iid]{
            \includegraphics[scale=0.40]{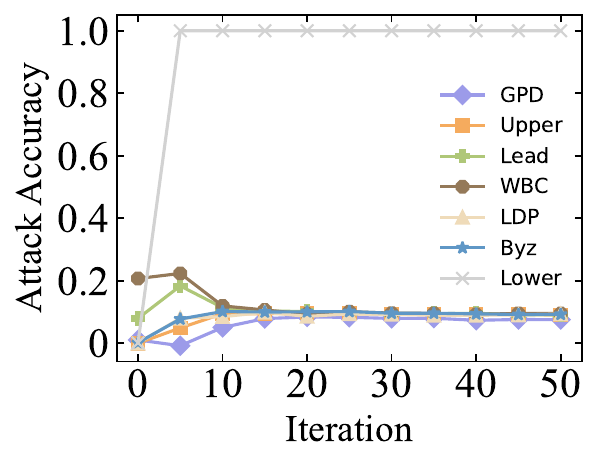}
        }
        \subfloat[\scriptsize non-overlap]{
            \includegraphics[scale=0.40]{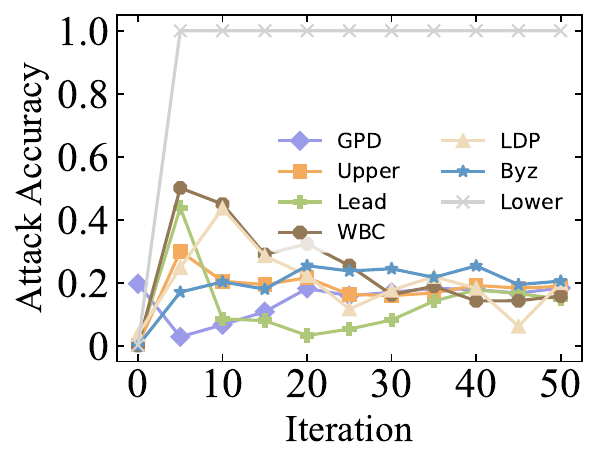}
        }
        \subfloat[\scriptsize label-dir]{
            \includegraphics[scale=0.40]{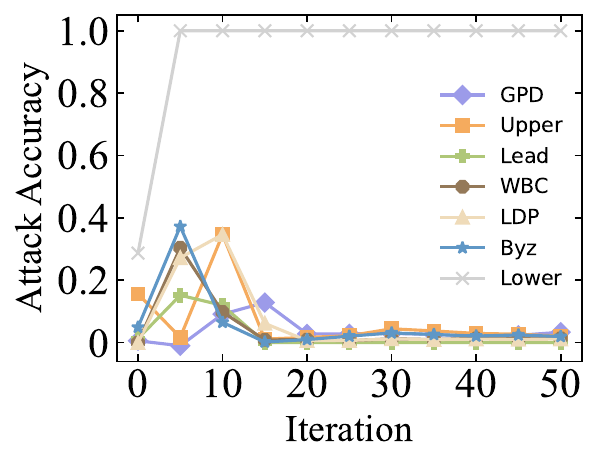}
        }
        \subfloat[\scriptsize quantity-dir]{
            \includegraphics[scale=0.40]{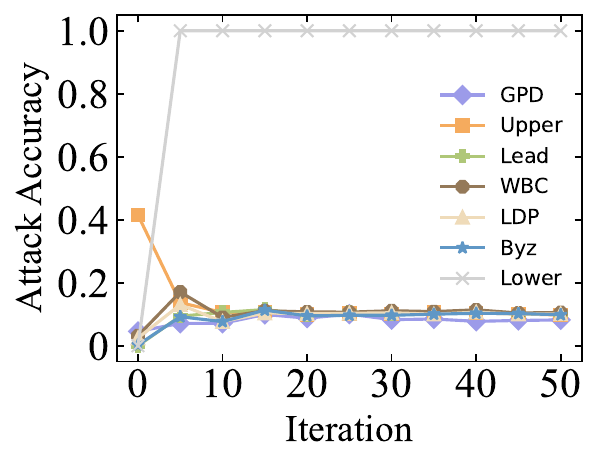}
        }
                        \vspace{-3pt}
        \caption{The attack accuracy among seven defenses and four data distributions under \emph{Backdoor-1-pixel} attack.}
                        \vspace{-5pt}
        \label{back-1}
    \end{minipage}

    \begin{minipage}{0.98\textwidth} 
    \captionsetup[subfloat]{font=footnotesize} 
        \subfloat[\scriptsize iid]{
            \includegraphics[scale=0.40]{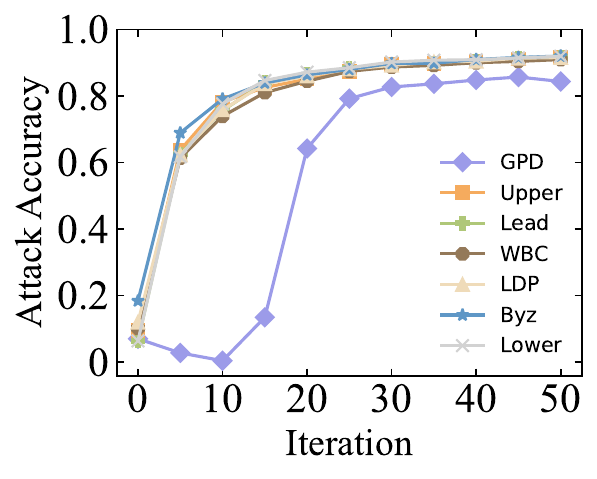}
        }
        \subfloat[\scriptsize non-overlap]{
            \includegraphics[scale=0.40]{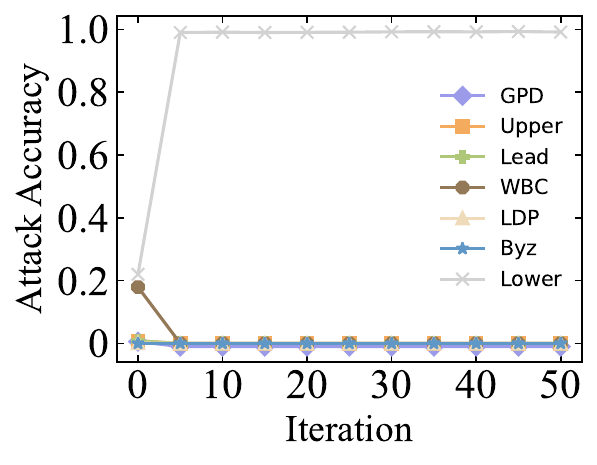}
        }
        \subfloat[\scriptsize label-dir]{
            \includegraphics[scale=0.40]{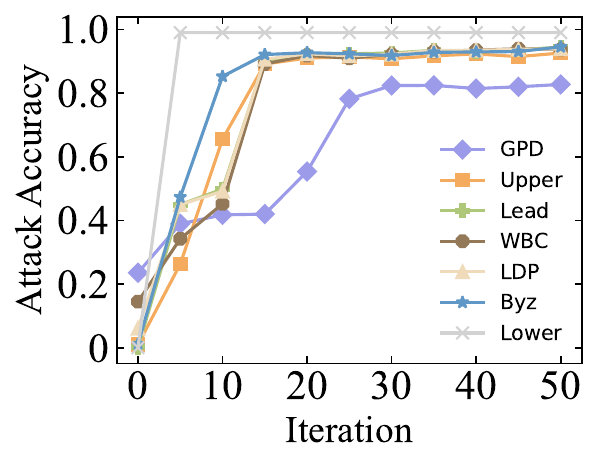}
        }
        \subfloat[\scriptsize quantity-dir]{
            \includegraphics[scale=0.40]{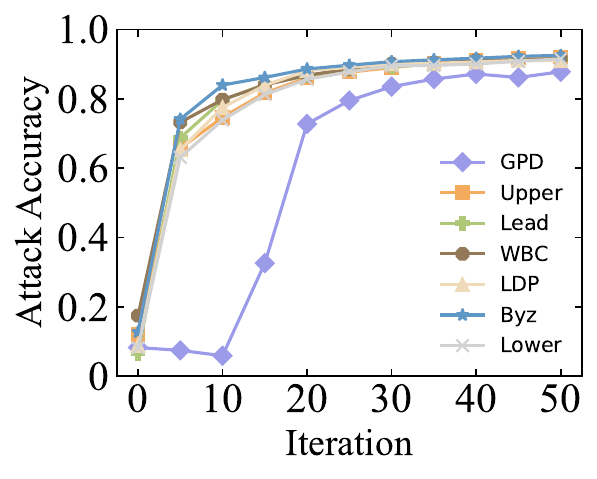}
        }
                        \vspace{-3pt}
    \caption{The attack accuracy among seven defenses and four data distributions under \emph{Single-image} attack.}
    \label{image-1}
    \end{minipage}
    \vspace{-8pt}
\end{figure*}

\begin{figure*}[t]
\centering
\vspace{-25pt}
    \begin{minipage}{0.98\textwidth} 
    \captionsetup[subfloat]{font=footnotesize} 
        \subfloat[\scriptsize iid]{
            \includegraphics[scale=0.40]{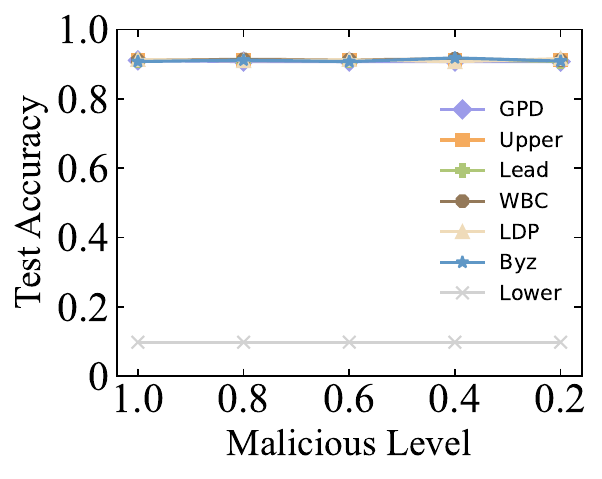}
        }
        \subfloat[\scriptsize non-overlap]{
            \includegraphics[scale=0.40]{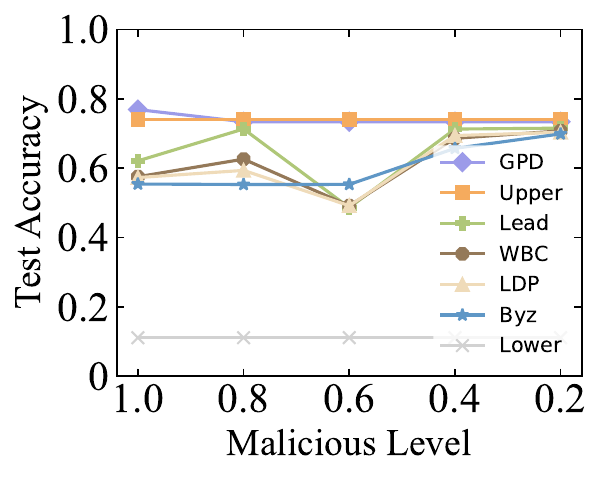}\label{level-3}
        }
        \subfloat[\scriptsize label-dir]{
            \includegraphics[scale=0.40]{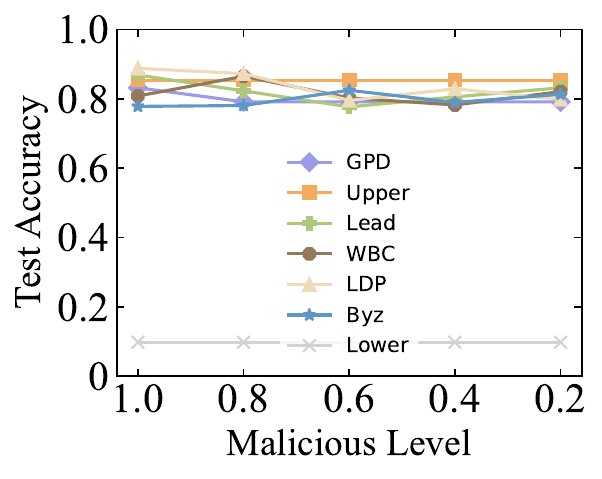}
        }
        \subfloat[\scriptsize quantity-dir]{
            \includegraphics[scale=0.40]{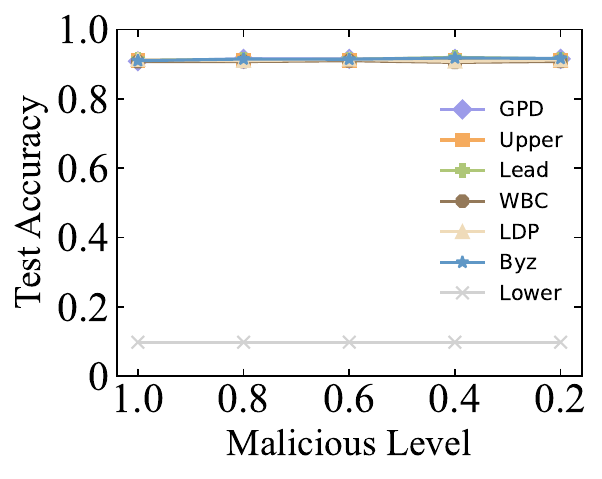}
        }
                        \vspace{-3pt}
    \caption{The test accuracy vs. \emph{malicious level}, among seven defenses and four data distributions under \emph{Backdoor-9-pixel} attack.}
   \vspace{-5pt}
    \label{level-1}
    \end{minipage}
    \begin{minipage}{0.98\textwidth} 
\captionsetup[subfloat]{font=footnotesize} 
        \subfloat[\scriptsize iid]{
            \includegraphics[scale=0.40]{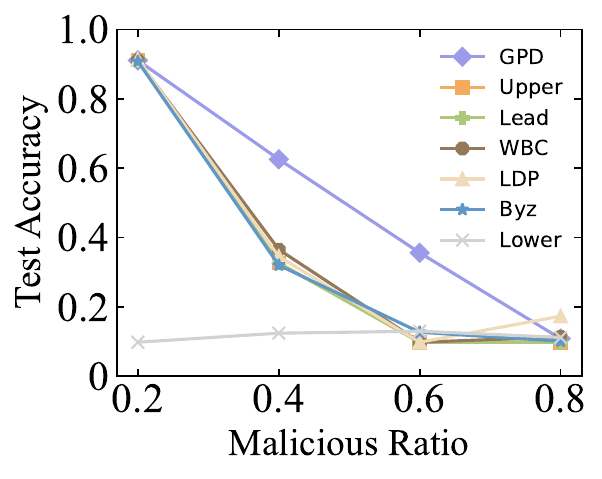}
        }
        \subfloat[\scriptsize non-overlap]{
            \includegraphics[scale=0.40]{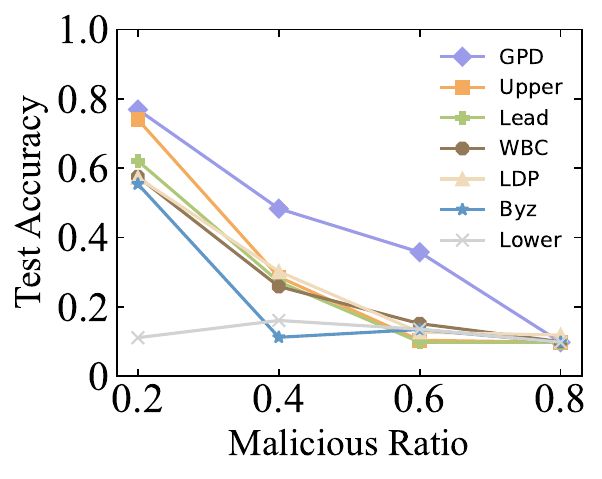}
        }
        \subfloat[\scriptsize label-dir]{
            \includegraphics[scale=0.40]{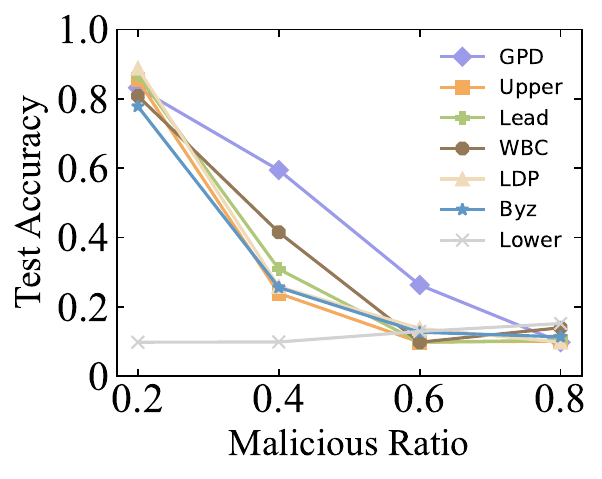}
        }
        \subfloat[\scriptsize quantity-dir]{
            \includegraphics[scale=0.40]{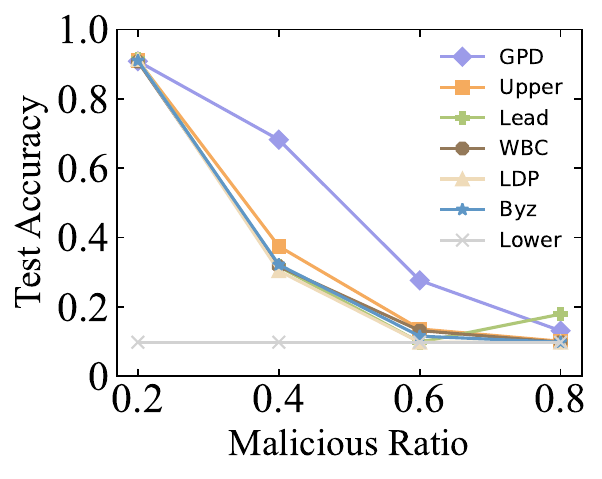}
        }
                        \vspace{-3pt}
    \caption{The test accuracy vs. \emph{malicious ratio}, among seven defenses and four data distributions under \emph{Backdoor-9-pixel} attack.}
    \label{ratio-1}
     \vspace{-5pt} 
    \end{minipage}

        \begin{minipage}{0.98\textwidth} 
\captionsetup[subfloat]{font=footnotesize} 
        \subfloat[\scriptsize iid]{
            \includegraphics[scale=0.40]{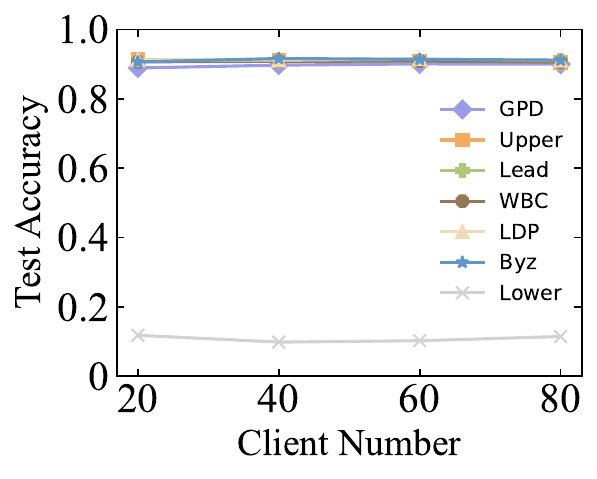}
        }
        \subfloat[\scriptsize non-overlap]{
            \includegraphics[scale=0.40]{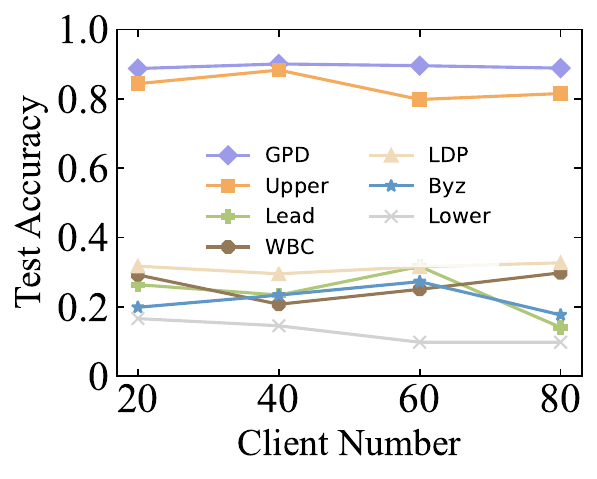}
        }
        \subfloat[\scriptsize label-dir]{
            \includegraphics[scale=0.40]{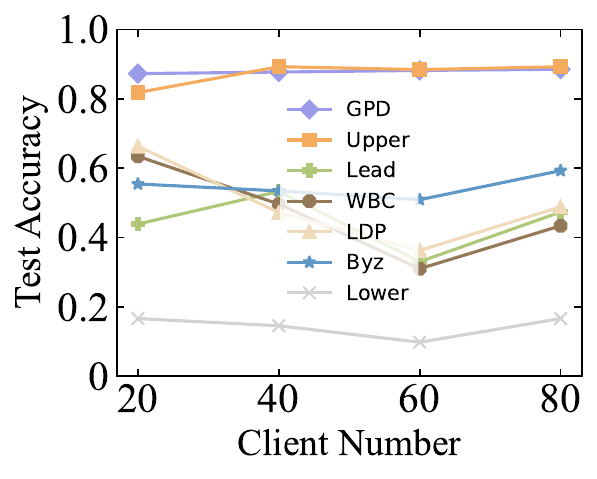}
        }
        \subfloat[\scriptsize quantity-dir]{
            \includegraphics[scale=0.40]{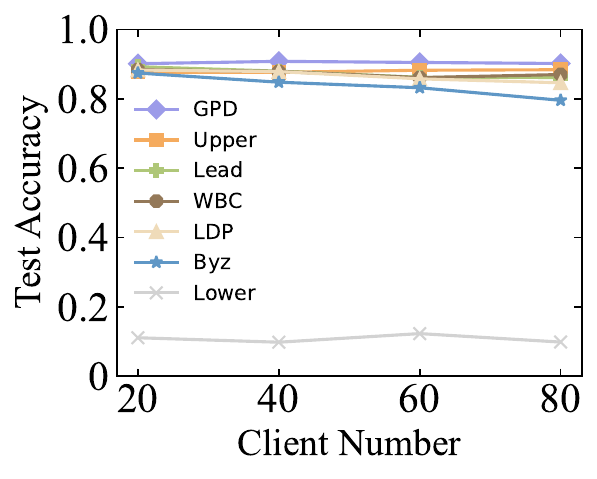}
        }
                        \vspace{-3pt}
    \caption{The test accuracy vs. \emph{client number}, among seven defenses and four data distributions under \emph{Backdoor-9-pixel} attack.}
    \label{number-1}
     \vspace{-5pt} 
    \end{minipage}

            \begin{minipage}{0.98\textwidth} 
\captionsetup[subfloat]{font=footnotesize} 
        \subfloat[\scriptsize iid]{
            \includegraphics[scale=0.40]{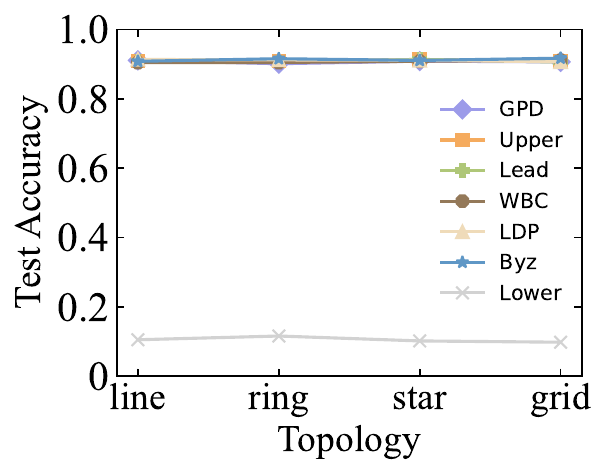}
        }
        \subfloat[\scriptsize non-overlap]{
            \includegraphics[scale=0.40]{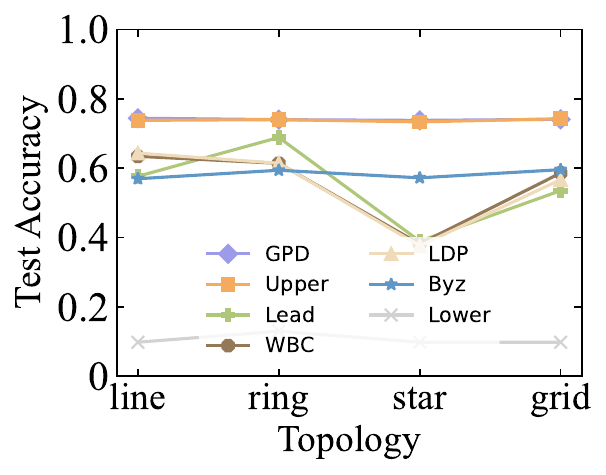}
        }
        \subfloat[\scriptsize label-dir]{
            \includegraphics[scale=0.40]{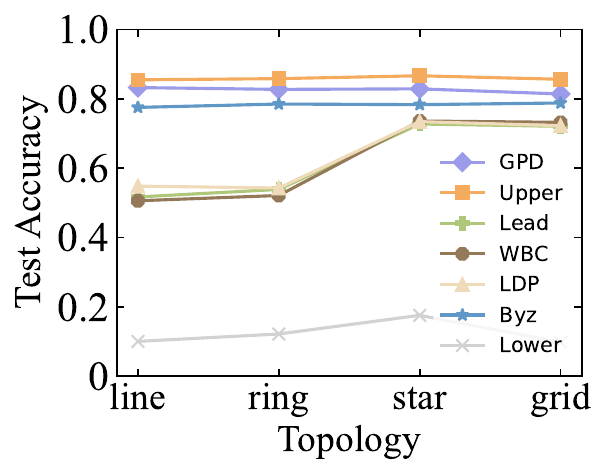}
        }
        \subfloat[\scriptsize quantity-dir]{
            \includegraphics[scale=0.40]{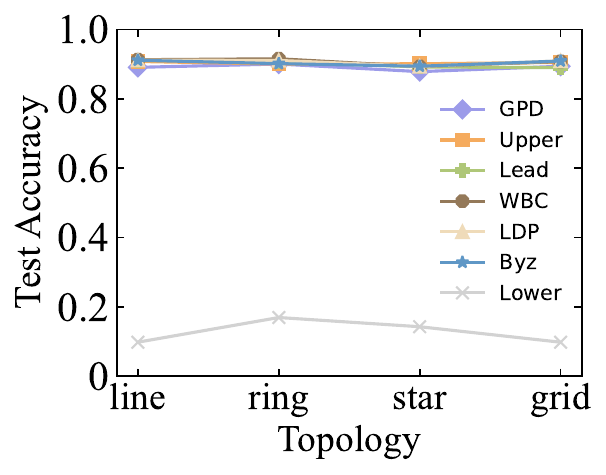}
        }
                        \vspace{-3pt}
    \caption{The test accuracy vs. \emph{topology}, among seven defenses and four data distributions under \emph{Backdoor-9-pixel} attack.}
    \label{topology-1}
     \vspace{-8pt} 
    \end{minipage}
\end{figure*}
\subsection{Comparison on Test Accuracy}
In the first set of experiments, we first plot the test accuracy curves for seven defenses and four data distributions, under \emph{Backdoor-9-pixel} attack and \emph{Lie} attack on the \textit{MNIST} datasets, as illustrated in Fig. \ref{back-9-acc} and Fig. \ref{lie-acc}. We also compare the average test accuracy across all attacks for different datasets: \textit{MNIST} and \textit{FashionMNIST} in Fig. \ref{data-1}, and \textit{CIFAR-10} and \textit{20Newsgroups} in Fig. \ref{data-2}. Notably, \textsf{GPD} demonstrates the highest model accuracy in terms of multiple data poisoning attacks, varying data distributions, and four datasets.

First, from a horizontal perspective, we evaluate the efficacy of various defense methods against \emph{multiple data poisoning attacks}. Notably, \textsf{GPD} consistently achieves the highest test accuracy across all data poisoning attacks, indicating it as the most effective defense. 
This is due to the fact that \textsf{GPD} enables benign clients to precisely mitigate all malicious gradients via recording variables, and optimize the model weights via purified gradients after mitigation. 
This process ensures the retention of previously beneficial contributions from malicious clients, thereby substantially enhancing the global model accuracy. In contrast, other methods such as \emph{LDP}, \emph{WBC}, \emph{Lead} and \emph{Byz} accidentally injure certain benign gradients and diminish the model accuracy. 
\textsf{GPD} achieves even better model accuracy than the \emph{Upper} method, which represents the upper bound of detect-restart defense methods. 
This is because the \emph{Upper} method lacks the capability to employ beneficial components within malicious gradients, thereby hindering the aggregated global model from achieving the optimal accuracy.

Second, from a vertical perspective, we assess the effectiveness of different defenses under \emph{varying data distributions}. It is evident that test accuracy consistently decreases when the label distribution is skewed (i.e., non-overlap and label-dir), especially in scenarios where client labels do not overlap (i.e., non-overlap). Nevertheless, \textsf{GPD} consistently achieves the highest accuracy across all data distributions.
This is due to its ability to maintain an unaffected global gradient aggregation, as indicated in Theorem \ref{theo-1}, which effectively counterbalances the influence of heterogeneous data distributions \cite{DBLP:conf/nips/AketiH023}. 
In contrast, other methods fail to maintain global gradient tracking and significantly hinder consistent convergence across clients, due to the introduction of random noise and regularization. This issue is particularly pronounced in the non-overlap case, where global gradient tracking and consistent convergence are more crucial for guiding clients to collaboratively train the global model. As a result, other methods experience aggregation failures, leading to a significant degradation in global model accuracy.

\vspace{-1em}
\subsection{Comparison on Attack Accuracy}
\vspace{-6pt}
In the second set of experiments, we first plot the attack accuracy curves for seven defenses and four data distributions under three data poisoning attacks on the \textit{MNIST} datasets, as depicted in Fig. \ref{back-9}, Fig. \ref{back-1}, and Fig. \ref{image-1}. 
The remaining three types of poisoning attacks are untargeted poisoning attacks and thus cannot report the attack accuracy.

First, from a horizontal perspective, we evaluate the attack accuracy across multiple data poisoning scenarios. As shown in Fig. \ref{back-9} and Fig. \ref{back-1}, the backdoor attack accuracy (i.e., backdoor accuracy) remains similar across different methods, including \textsf{GPD}. 
This similarity stems from the uniform adoption of the same consistency-based detection algorithm for malicious client identification. 
The backdoor accuracy metric primarily reflects the detection effectiveness, i.e., the algorithm's capability to identify and exclude malicious clients. Consequently, the identical detection naturally leads to similar backdoor accuracy results.
Furthermore, empirical analysis reveals that single-image attacks achieve higher attack accuracy than backdoor attacks. This phenomenon can be attributed to the fact that single-image attacks focus on manipulation of individual class labels while preserving the integrity of other class distributions. 
This characteristic subsequently creates circumstances wherein malicious clients remain difficult to detect during prior stages, allowing harmful impact to persist in baseline algorithms. In contrast, \textsf{GPD} effectively mitigates all malicious gradients from the beginning once detection occurs through its recording variable mechanism. Consequently, it attains substantially lower attack accuracy, signifying enhanced defensive capabilities.

Second, from a vertical perspective, we assess attack accuracy under \emph{varying data distributions}.
It is evident that each defense method exhibits significant fluctuations under label distribution skew. Moreover, \textsf{GPD} consistently demonstrates low attack accuracy, underscoring its robustness against poisoning attacks.
This advantage stems from the fact that the recording variables in \textsf{GPD} can accurately track historically aggregated gradient tracking variables, thus effectively mitigating the malicious impact after detection. 
In contrast, other methods that introduce random noise or regularization per iteration are unable to steadily counteract malicious impact, especially in heterogeneous data distributions.

\begin{figure}[!t]
    \centering
   \vspace{-10pt}
  \captionsetup[subfloat]{font=footnotesize} 
    \hspace{-3pt}
    \subfloat[\scriptsize Line]{
        \includegraphics[scale=0.12]{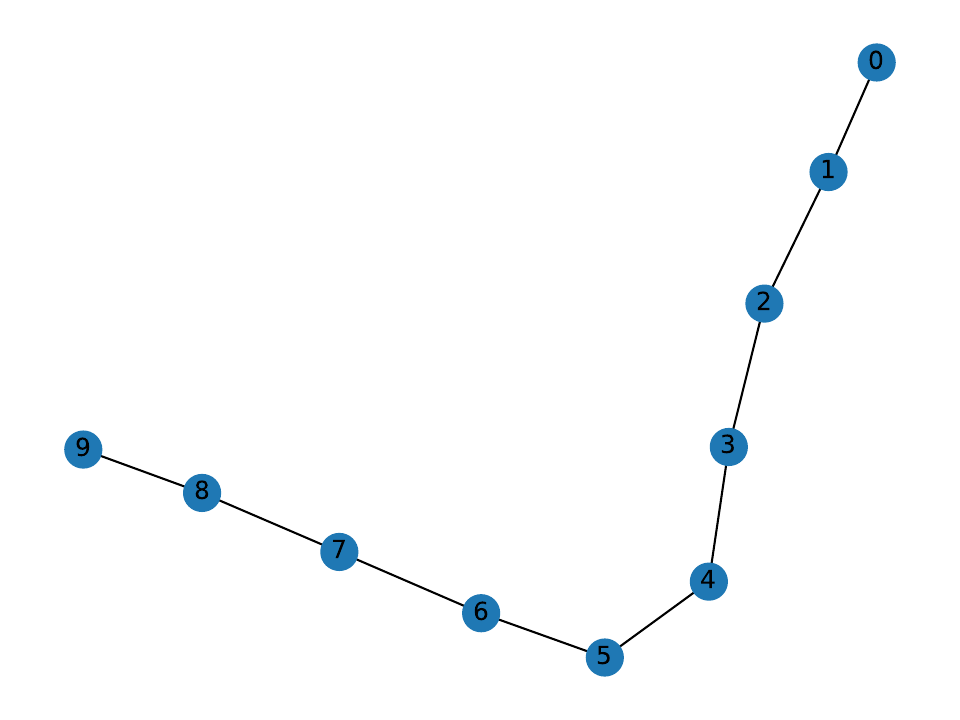}
    }
        \subfloat[\scriptsize Ring]{
        \includegraphics[scale=0.12]{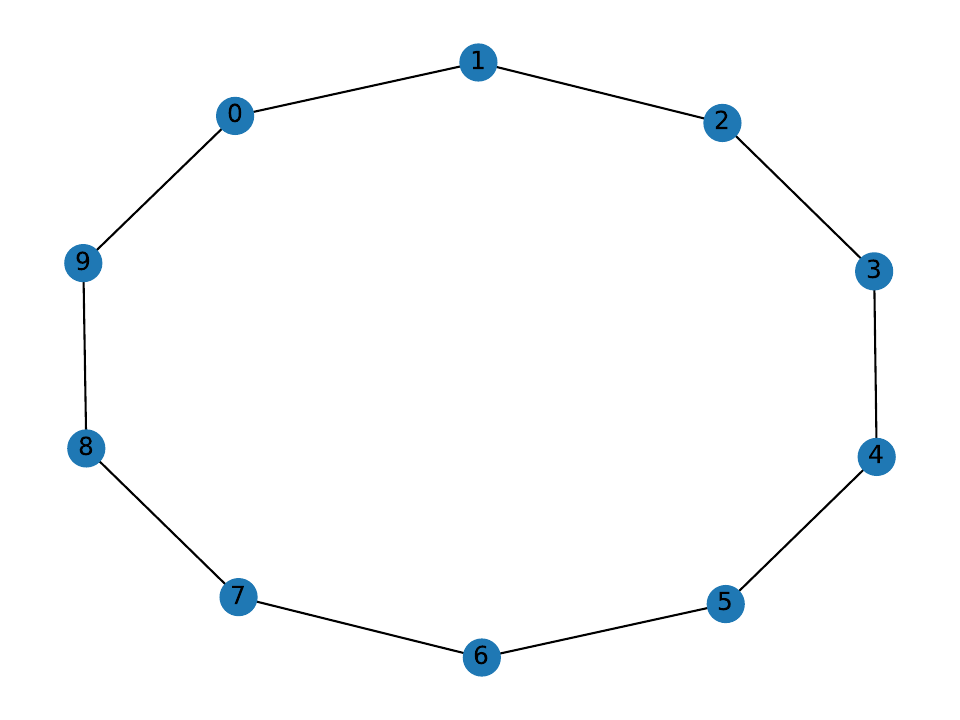}
    }
        \subfloat[\scriptsize Star]{
        \includegraphics[scale=0.12]{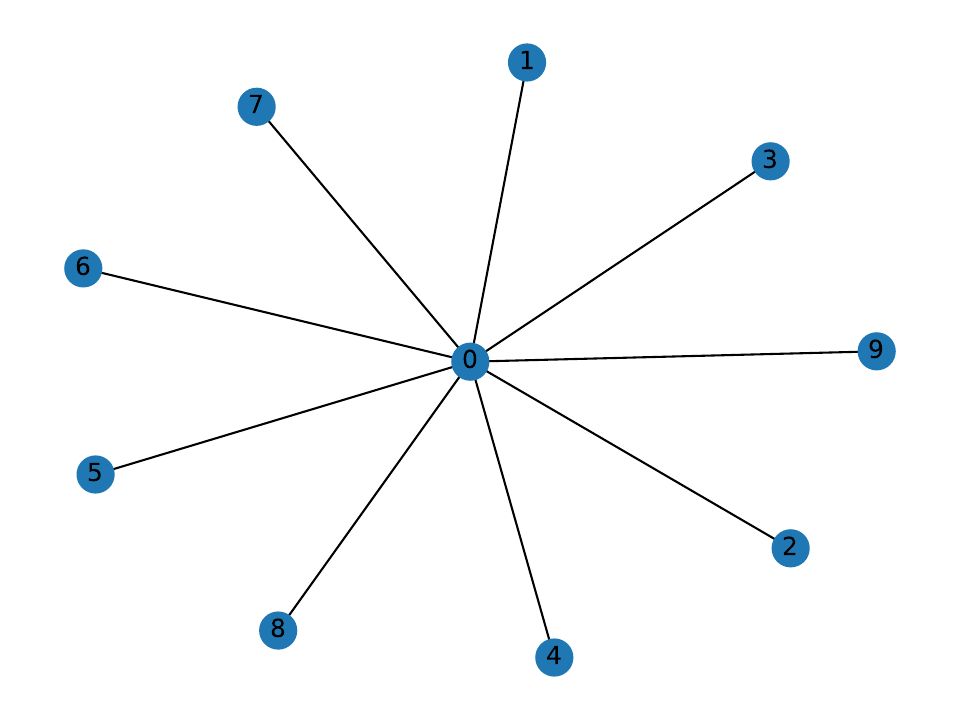}
    }
        \subfloat[\scriptsize Grid]{
        \includegraphics[scale=0.12]{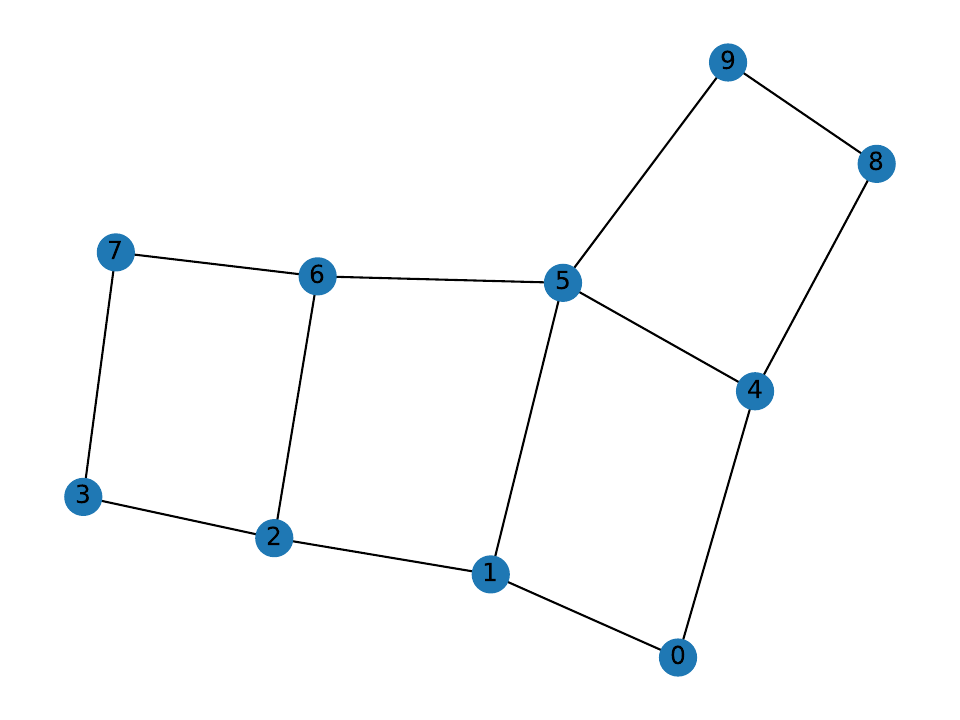}
    }
    \caption{The communication topology: schematic diagram.}
    \label{topology-vis}
    \vspace{-8pt}
\end{figure}

\subsection{Parameter Evaluation}

\emph{Effect of malicious level.} When varying the malicious level $\pi$ from 1.0 to 0.2, the corresponding experimental results under \emph{Backdoor-9-pixel} backdoor attack over the \textit{MNIST} dataset are presented in Fig. \ref{level-1}. Similar trends are observed for other types of attacks. 
The results indicate that as the malicious level decreases, the model accuracy across various defenses generally remains stable. This is because the detection algorithm gradually reduces the aggregation weight of malicious clients; even when the malicious level of a malicious client is minimal, the \emph{cumulative} malicious impact from previous iterations can still be detected by benign clients.
However, in the non-overlap case, where client data labels are entirely inconsistent, methods such as \emph{LDP} and \emph{WBC} struggle to maintain global gradient tracking, making DFL aggregation more vulnerable to disruption. As shown in Fig. \ref{level-3}, in the non-overlap case, these methods fail to effectively detect and mitigate malicious gradients, leading to a significant fluctuation in the global model accuracy.

\emph{Effect of malicious ratio.} Varying the malicious ratio (i.e., the proportion of malicious clients among all clients) from 0.2 to 0.8, the corresponding experimental results under the \emph{Backdoor-9-pixel} attack on the \emph{MNIST} dataset are depicted in Fig. \ref{ratio-1}. Similar patterns are observed for other attack types. Notably, as the ratio of malicious clients escalates, the cumulative effect of adversarial behavior grows increasingly dominant, overwhelming existing defense mechanisms' capacity to reliably identify malicious clients. This systemic vulnerability precipitates comprehensive defense collapse, accompanied by substantial model accuracy degradation. In the case of the \emph{Upper} method, progressive increases in the malicious client ratio exacerbate data scarcity among benign clients, particularly under label distribution skew scenarios, thereby also compounding accuracy degradation.
Remarkably, our proposed \textsf{GPD} framework consistently achieves superior test accuracy compared to other approaches. This performance advantage reaffirms its efficacy in both enhancing model accuracy and optimally leveraging available data assets.

\emph{Effect of client number.} Fig. \ref{number-1} presents the experimental results when varying the number of clients from 20 to 80 under the \emph{Backdoor-9-pixel} attack on the \emph{MNIST} dataset. As observed, the test accuracy remains relatively stable across different client numbers for the proposed \textsf{GPD} method. Moreover, in the non-overlap and label-dir scenarios, increasing the number of clients introduces greater heterogeneity in data distribution, posing additional challenges for other baseline methods, i.e., huge accuracy degradation. \textsf{GPD} maintains consistent performance across varying client numbers, demonstrating its robustness to network scale variations.

\begin{table}[t]
\centering
\caption{\centering \color{black} Test accuracy ($\uparrow$) under different scale parameters (\%).}
\vspace{5pt}
\label{alpha}
\setlength{\tabcolsep}{6pt}
\begin{tabular}{cccccccc}
\shline
  $\alpha$          & GPD & Upper & Lead & WBC & LDP &Byz &Lower \\ \hline
 0.2 & \textbf{83.49} & 82.81 & 74.69 & 65.89 & 74.65 & 75.49 & 12.74 \\ \hline
 0.4 & 90.39 & \textbf{90.73} & 83.34 & 87.60 & 82.36 & 85.76 & 10.35 \\ \hline
 0.6 & \textbf{89.77} & 89.70 & 87.35 & 88.30 & 89.33 & 84.61 & 11.05 \\ \hline
 0.8 & \textbf{89.62} & 88.52 & 87.02 & 87.36 & 86.41 & 89.32 & 10.04 \\ \hline
 1.0 & 89.66 & \textbf{89.87} & 88.49 & 88.73 & 86.67 & 86.76 & 9.74 \\ \shline
\end{tabular}
\vspace{-8pt}
\end{table}

\emph{Effect of topology.} The impact of different network topologies (line, ring, star, and grid) on defense performance is illustrated in Fig. \ref{topology-1}. Moreover, the schematic diagrams of communication topologies are plotted in Fig. \ref{topology-vis}.
As shown in Fig. \ref{topology-1}, \textsf{GPD} exhibits remarkable consistency across all topology configurations, particularly in maintaining test accuracy under the challenging label-distribution skew scenario (non-overlap and label-dir). This topology-invariant performance underscores the adaptability of our proposed GPD to diverse communication topologies.

\emph{Effect of scale parameter.} To evaluate the robustness of defense mechanisms under varying degrees of data heterogeneity, we conduct experiments with different Dirichlet distribution scale parameters $\alpha$ ranging from 0.2 to 1.0, as reported in Table \ref{alpha}. 
We tested different scale parameters in label distribution skew (i.e., label-dir), which provides greater influence than quantity distribution skew \cite{li2022federated}.
The experimental results under the \emph{Backdoor-9-pixel} attack on the \emph{MNIST} dataset demonstrate the impact of data heterogeneity on defense performance. 
As $\alpha$ increases, indicating more balanced data distributions, the performance gap among different methods narrows, with all methods showing improved accuracy. However, \textsf{GPD} consistently demonstrates competitive or superior performance across all $\alpha$ values. This robust performance across varying degrees of data heterogeneity underscores the effectiveness of our proposed \textsf{GPD} method in handling heterogeneous data distributions.

\begin{table}[t]
\centering
\caption{\centering Test accuracy ($\uparrow$) on clean data label `1', attack accuracy ($\downarrow$) on poisoned data label `0' and overall test accuracy (\%).}
\vspace{5pt}
\label{tab:ablation_beneficial}
\setlength{\tabcolsep}{3.2pt}
\begin{tabular}{lccccccc}
\shline
          & \textsf{GPD} & Upper & Lead & WBC & LDP & Byz & Lower \\ \hline
          Test accuracy ($\uparrow$) & \textbf{28.35} & 0 & 4.80 &  1.15 & 1.40 & 0 & 99.48
\\ \hline
Attack accuracy ($\downarrow$) & 8.32 & 5.05 & 34.65 & 32.57 & 26.73 & 29.70 & 97.23 \\ \hline
Overall accuracy ($\uparrow$) & 75.43 & 73.99 & 56.83 & 54.72 & 57.60 & 57.88 & 22.45  \\ 
\shline
\end{tabular}
\vspace{5pt}
\end{table}
\begin{table}[t]
\centering
\caption{\centering Test accuracy ($\uparrow$) under different detection algorithms (\%).}
\vspace{5pt}
\label{ablat}
\setlength{\tabcolsep}{4.7pt}
\begin{tabular}{cccccccc}
\shline
Based           & GPD  & Upper & Lead & WBC & LDP &Byz &Lower \\ \hline
Consistency & \textbf{91.18} & 91.16    & 91.07  & 91.10 & 91.16 &91.24 & 10.09   \\ \hline
Similarity & \textbf{91.34}  & 91.16   & 90.68  & 90.75 & 91.04 &90.97 & 10.09   \\ \hline
None      & 10.09  & 91.16   & 10.09  & 10.09 & 10.09 &10.09 &10.09   \\ \shline
\end{tabular}
\vspace{-8pt}
\end{table}

\subsection{Ablation Study}
In this set of experiments, we first quantify the retained beneficial components from malicious clients, and then test the effectiveness of different detection algorithms.
\subsubsection{Quantifying Beneficial Components}

To rigorously quantify the contribution of these beneficial components, we design an ablation study where this contribution could be clearly isolated and measured.

We establish an experimental setting with an extreme non-overlap data distribution among five clients using the \emph{MNIST} dataset. Four benign clients respectively hold data from labels `2' to `9', while one malicious client exclusively possesses data for labels `0' and `1'. The information about labels `0' and `1' is entirely absent from the benign set. The malicious client executes a backdoor attack on data for \emph{label `0'} while maintaining \emph{clean data for label `1'}.

In this scenario, the information about label `1' represents the beneficial components unique to the malicious client. We quantify its contribution by measuring the final global model's test accuracy on this specific label `1', comparing \textsf{GPD} against the Upper method where the malicious client is completely excluded from DFL aggregation.

As Table \ref{tab:ablation_beneficial} demonstrates, the Upper method, having never encountered data for label `1', achieves 0\% test accuracy on it. Under the extreme non-overlap data distribution, other baseline methods fail to achieve the balance between retaining beneficial information and mitigating harmful information. In contrast, our \textsf{GPD} method successfully defends against the backdoor attack on label `0' (i.e., 8.32\% attack accuracy) while attaining 28.35\% test accuracy on label `1'. This provides quantifiable evidence that \textsf{GPD} effectively extracts and utilizes beneficial components from malicious clients, enhancing the global model's performance beyond what is achievable through simple client exclusion.

\subsubsection{Different Detection Algorithm}
From the above experiments, it can be observed that the effect of the detection algorithm is of crucial significance to the performance of defense methods. Hence, we compare the consistency-based detection, similarity-based detection, and none detection in this experiment, as shown in Table \ref{ablat}.

Table \ref{ablat} presents the test accuracy of seven defenses across different detection algorithms under the \emph{Backdoor-9-pixel} attack. It can be observed that all defenses are compatible with both consistency-based and similarity-based detection approaches. However, these methods cannot operate independently without detection algorithms, as their effectiveness in mitigating malicious impact depends entirely on the identification results generated by the detection algorithms.

It is important to note the inherent trade-off in detection algorithm speed and the retention of beneficial components.
A rapid, aggressive detection algorithm might exclude malicious clients early—excellent for security but limiting opportunities for encoding their beneficial knowledge into model weights. Conversely, a more gradual detection approach might allow malicious clients to participate longer, providing more time for beneficial information encoding but introducing significant risks: prolonged activity of malicious clients can compromise the detection mechanism itself, potentially resulting in misclassification of benign clients and detection failure.

\begin{table}[t]
\centering
\caption{\centering Computation time ($\downarrow$) under different detection algorithms (s).}
\vspace{5pt}
\label{time-1}
\setlength{\tabcolsep}{4pt}
\begin{tabular}{cccccccc}
\shline
   & GPD & Upper & Lead & WBC & LDP &Byz &Lower \\ \hline
 Time(s)   & 0.1051 & 0.0913 & 0.1042 & 0.1182 & 0.1142 & 0.1131 & 0.1133   \\ \shline
\end{tabular}
\vspace{-8pt}
\end{table}

\subsection{Overhead Analysis}
Finally, we assess the computational and resource overhead associated with \textsf{GPD} in comparison to other defense methods. (i) The recording variables are maintained exclusively within local environments and require no inter-client transmission, thereby introducing \emph{no} additional communication overhead. (ii) Each client is required to maintain only $|\mathcal{N}_i|$ recording variables in local storage, imposing minimal memory requirements. Here, $|\mathcal{N}_i|$ represents the number of neighboring clients for client $c_i$. (iii) The computational overhead involved in processing the recording variables is negligible, necessitating merely a single matrix operation per iteration. 
For fair comparison, we measure the end-to-end execution time of all baseline algorithms, explicitly accounting for the computational overhead introduced by the consistency-based detection algorithms.
As evidenced in Table \ref{time-1}, which presents the per-iteration computation time for each client, \textsf{GPD} demonstrates computational efficiency comparable to existing defense methods while simultaneously achieving superior performance.

\section{Conclusion}
\label{section-7}
To defend against data poisoning attacks and enhance model performance within decentralized federated learning (DFL), we introduce a novel gradient purification defense, \textsf{GPD}, to mitigate the malicious impact and enhance model accuracy. 
Each benign client in \textsf{GPD} maintains the recording variable to track all historically aggregated gradients from each neighbor. Via historical consistency checks, benign clients can detect malicious neighbors and mitigate all recorded malicious gradients. Then model weights are optimized by purified gradients, which skillfully retain the beneficial contributions from malicious clients. It significantly enhances the model accuracy and we prove the ability of \textsf{GPD} to harvest high accuracy. Empirical evaluations demonstrate that our solution significantly outperforms state-of-the-art defenses in terms of model accuracy against various data poisoning attacks. Moreover, \textsf{GPD} significantly performs better than other methods in various heterogeneous data distributions.


\section*{Acknowledgments}
This work is supported by Leading Goose R\&D Program of Zhejiang (No. 2024C01109), the NSFC (No. 62372404), and Fundamental Research Funds for Central Universities \\(No. 226-2024-00030). Xiaoye Miao is the corresponding author of this work.

\balance
\bibliographystyle{fcs}
\bibliography{ref}





\end{document}